\pdfoutput=1

\documentclass[11pt]{article}

\usepackage[final]{acl}
\usepackage{times}
\usepackage{latexsym}
\usepackage{lipsum}
\usepackage{booktabs}
\usepackage[T1]{fontenc}
\usepackage[utf8]{inputenc}
\usepackage{microtype}
\usepackage{inconsolata}
\usepackage{graphicx}
\usepackage{multirow}
\usepackage{amsmath}
\usepackage{algorithm}
\usepackage{algorithmic}
\usepackage{amsfonts}
\usepackage[dvipsnames]{xcolor}

\title{Seeing Culture: A Benchmark for Visual Reasoning and Grounding}

\author{
 \textbf{Burak Satar\textsuperscript{1*}},
 \textbf{Zhixin Ma\textsuperscript{1*}},
 \textbf{Patrick A. Irawan\textsuperscript{2}},
 \textbf{Wilfried A. Mulyawan\textsuperscript{2}},
\\
 \textbf{Jing Jiang\textsuperscript{1}},
 \textbf{Ee-Peng Lim\textsuperscript{1}},
 \textbf{Chong-Wah Ngo\textsuperscript{1}},
\\
\texttt{\{buraks, zxma, jingjiang, eplim, cwngo\}@smu.edu.sg}
\\
\texttt{\{patrickamadeusirawan, arielwilfried0\}@gmail.com}
\\
 \textsuperscript{1}Singapore Management University,
 \textsuperscript{2}Bandung Institute of Technology
\\\\
\includegraphics[height=1.75ex]{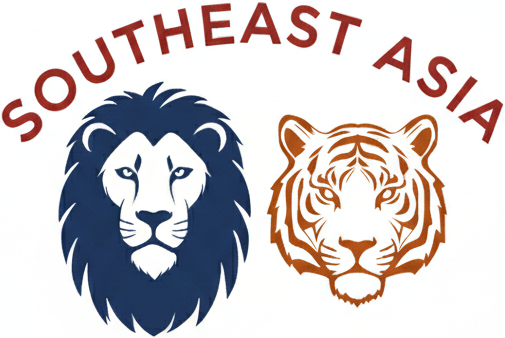}   {\color{NavyBlue} https://seeingculture-benchmark.github.io} \\
\includegraphics[height=1.75ex]{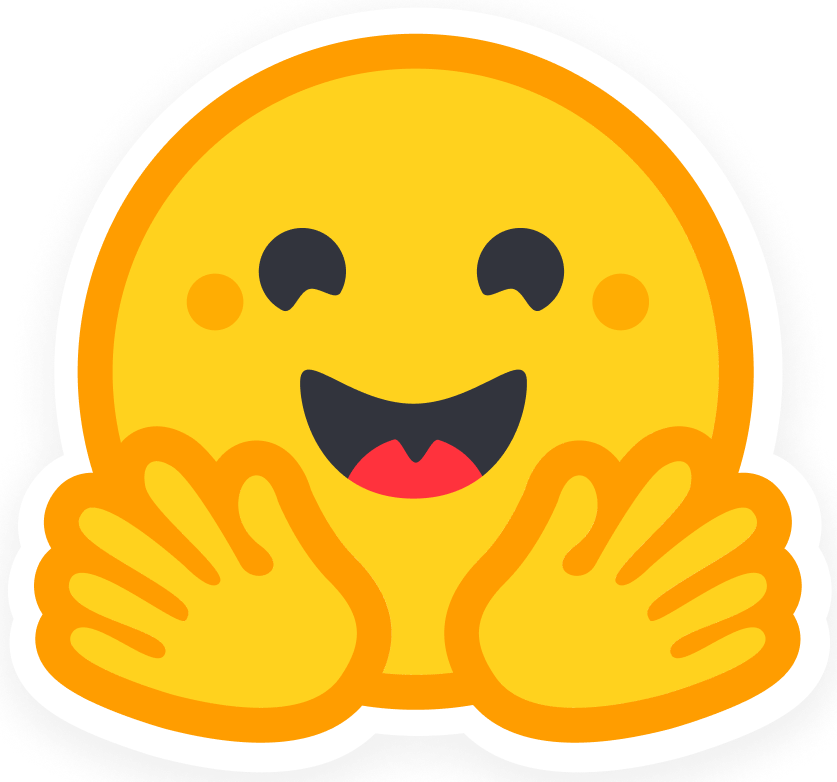}   {\color{NavyBlue} https://huggingface.co/datasets/Multimedia-SMU/seeingculture-benchmark}\\
}

\begin{document}
\maketitle

\begin{abstract}
Multimodal vision-language models (VLMs) have made substantial progress in various tasks that require a combined understanding of visual and textual content, particularly in cultural understanding tasks, with the emergence of new cultural datasets. However, these datasets frequently fall short of providing cultural reasoning while underrepresenting many cultures.
In this paper, we introduce the Seeing Culture Benchmark (SCB), focusing on cultural reasoning with a novel approach that requires VLMs to reason on culturally rich images in two stages: i) selecting the correct visual option with multiple-choice visual question answering (VQA), and ii) segmenting the relevant cultural artifact as evidence of reasoning. Visual options in the first stage are systematically organized into three types: those originating from the same country, those from different countries, or a mixed group. Notably, all options are derived from a singular category for each type. Progression to the second stage occurs only after a correct visual option is chosen. 
The SCB benchmark comprises 1,065 images that capture 138 cultural artifacts across five categories from seven Southeast Asia countries, whose diverse cultures are often overlooked, accompanied by 3,178 questions, of which 1,093 are unique and meticulously curated by human annotators. 
Our evaluation of various VLMs reveals the complexities involved in cross-modal cultural reasoning and highlights the disparity between visual reasoning and spatial grounding in culturally nuanced scenarios. The SCB serves as a crucial benchmark for identifying these shortcomings, thereby guiding future developments in the field of cultural reasoning. \\\includegraphics[height=1.5ex]{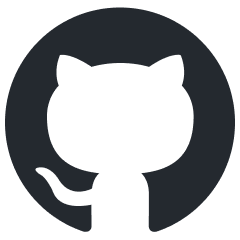} {\color{NavyBlue}https://github.com/buraksatar/SeeingCulture}
\end{abstract}

\section{Introduction}
\label{sec:intro}

\begin{figure}
  \includegraphics[width=1\linewidth]{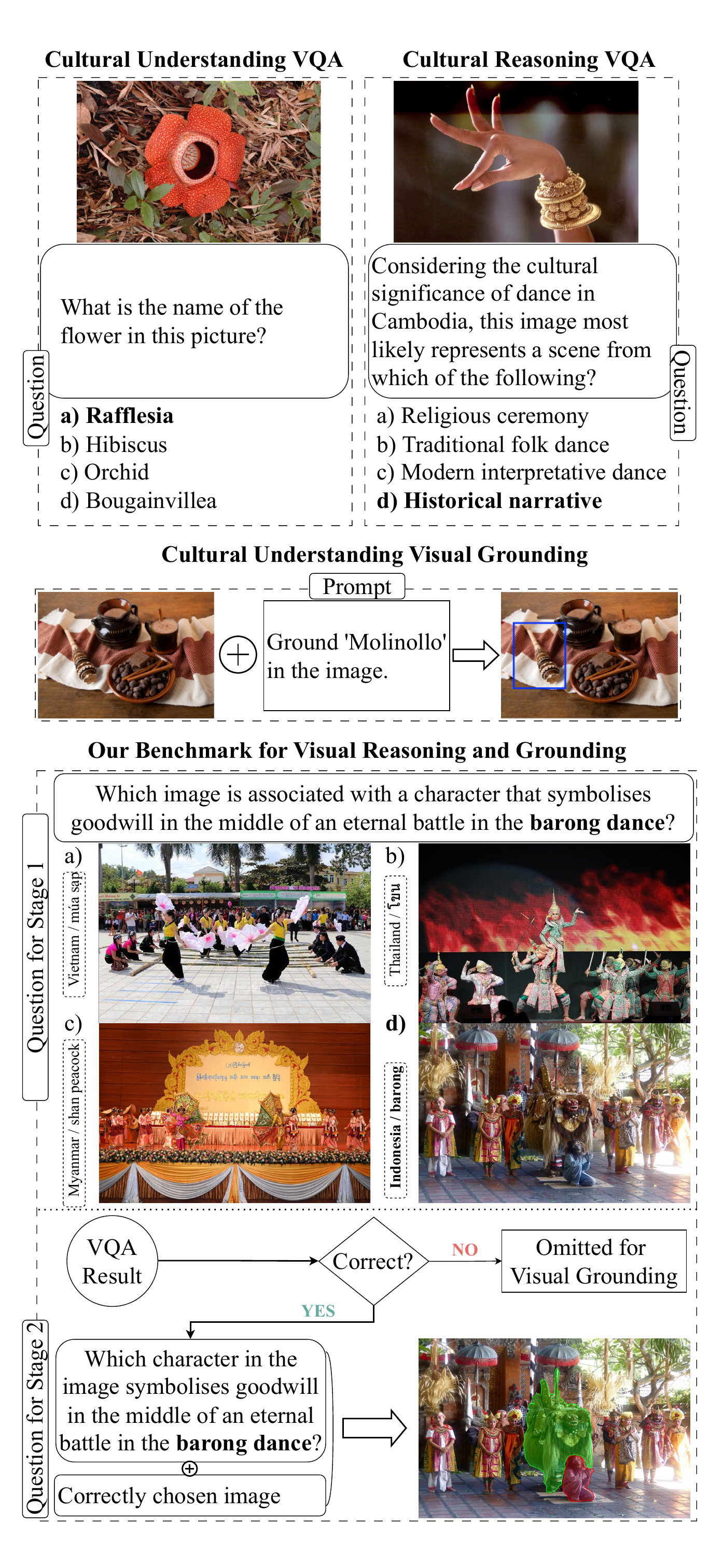}
  \caption{Comparison between our benchmark (SCB) and the recent studies on cultural understanding \cite{mogrovejo2024cvqa, bhatia-etal-2024-local_GlobalRG} and reasoning \cite{2024_seavqa}. SCB requires reasoning on cultural artifacts via diverse and rich visuals.}
  \label{fig:intro}
\end{figure}

\begin{figure*}[!htb]
  \includegraphics[width=1\textwidth]{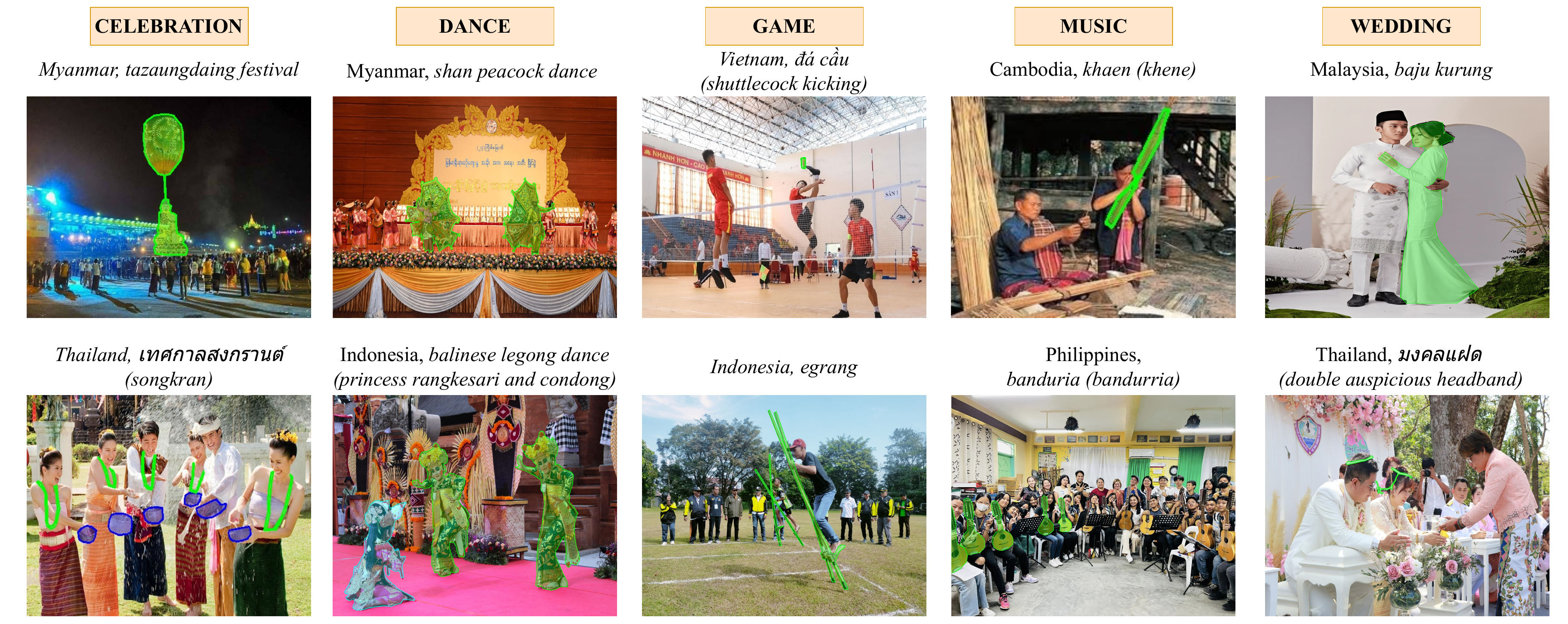}
  \caption{The presented collection of images from our SCB encompasses visual representations of cultural concepts from seven countries, categorized across five dimensions: music, game, dance, celebration, and wedding. These images exhibit either a variety of cultural artifacts situated in diverse contexts (e.g., the depiction of the \textit{balinese legong dance} showcases multiple characters, two \textit{princesses rangkesari}, and one \textit{condong}, with corresponding questions) or integrated distractors in addition to the primary concept (e.g., the image featuring the \textit{banduria}, which displays Spanish guitars on the right side while the \textit{bandurias} are positioned on the left). The segmentation masks of concepts are best viewed in color.}
  \label{fig:sample}
\end{figure*}

{\def\thefootnote{}\footnotetext{*Equal contribution.} Recent multimodal VLMs have demonstrated impressive performance on various tasks, such as VQA and visual grounding, which require assessing the understanding of both visual and textual information. For instance, VQA tasks with open-ended or multiple-choice questions have been used on various generic topics such as healthcare and entertainment. At the same time, visual grounding, which entails segmenting an object based on textual input, has predominantly expanded general scene understanding via recent VLMs. However, their performance may vary significantly across different cultural contexts, underscoring the need for new benchmarks to assess and enhance their performance in diverse cultural contexts. While recent studies \cite{nayak-etal-2024-benchmarking_CulturalVQA, Wang_CVLUE_2025, mogrovejo2024cvqa, bhatia-etal-2024-local_GlobalRG} attempt to address this gap with a focus on cultural understanding, there remains a pressing need for more comprehensive datasets that encompass a wider array of cultural nuances and artifacts, ensuring that VLMs can reason on culturally specific queries. We must emphasize that cultural reasoning involves not only recognizing cultural artifacts but also understanding their significance within specific contexts. For instance, considering our example in Figure \ref{fig:intro}, certain clues need to be taken into account, such as the fact that the \textit{barong dance} belongs to a specific culture, which differentiates it from other visual options, as well as the various characters that symbolize different meanings.
Creating such adequate benchmarks for cultural reasoning is challenging due to the various factors that influence cultural representation, such as the selection of images, the formulation of questions, and the data collection process. Despite providing essential insights, the present benchmarks exhibit significant limitations. For instance, \cite{2024_seavqa, corr_kviscuit, liu2025culturevlm, schneider2025gimmickgloballyinclusive} focus on cultural reasoning VQA; however, many of the images lack distractors, focusing solely on the cultural concept, while the questions are AI-generated, which may lack authenticity in cultural representation. Additionally, textual answers to the traditional VQA approaches may be influenced by spurious correlations \cite{fu-etal-2023-generate, Liu_Fan_Zhou_Xu_2023, 10214252, wang-etal-2023-dataset, Zhang_Zhang_Xu_2024} regardless of their design, as addressed by recent works. Furthermore, benchmarks specific to the segmentation task in this context have yet to be developed.       

To this end, we propose the Seeing Culture Benchmark, a novel benchmark to assess the cultural reasoning of VLMs in Southeast Asia countries, providing diversity in culture, given their limited resources in cultural representation within existing datasets. SCB includes complex images with rich and varied cultural contexts, paired with thoughtfully crafted questions that challenge the model's understanding and reasoning of cultural specifics in two stages: i) The multiple-choice options contain images representing diverse cultural artifacts, ii) The segmentation of cultural artifacts plays a role as evidence of reasoning. Advancement to the subsequent stage takes place only by following an accurate visual selection. Moreover, we ensure that the questions can reflect authentic cultural narratives through two rounds of verification with native speakers and cultural experts. This human-centric approach, which bypasses AI, avoids potential biases in content creation. Thus, our approach provides a more holistic view of the context, requiring VLMs to reason about the relationships between different cultural elements, thereby enhancing the depth of cultural reasoning. Our benchmark comprises five main categories, 138 cultural concepts, 1,065 images, and 3,178 questions from seven Southeast Asian countries, as depicted in Figure \ref{fig:sample}.

Further, we systematically evaluate several state-of-the-art VLMs on three distinct types. Type 1 consists of options originating from the same country, while Type 2 encompasses options from different countries in relation to the correct answer. Type 3 consists of a blend of Type 1 and Type 2 options. The sole commonality among these types is category consistency for all options (e.g., dance). The results indicate that VLMs perform the least on Type 1 questions, display the highest performance on Type 2 questions, and exhibit intermediate performance on Type 3 questions. This suggests that cues within the questions regarding the country or specific regional cultures can aid in discerning the correct answer. Moreover, there is a notable discrepancy between visual reasoning and spatial grounding, suggesting that although VLMs may select the correct option, they frequently lack the capacity to substantiate their reasoning through grounding. Consequently, the SCB is vital for fostering cross-modal reasoning in a culturally sensitive framework, shedding light on the disparity between visual reasoning and grounding. Our research will aid in developing more culturally conscious models, thereby improving their functionality in reasoning across diverse cultural contexts.

\begin{table*}[!htb]
\resizebox{\textwidth}{!}{%
\begin{tabular}{|c|c|c|c|c|c|c|c|c|c|c|c|}
\hline
\textbf{Dataset}                                                                      & \textbf{Country} & \textbf{Category} & \textbf{Concept} & \textbf{Image} & \textbf{Question} & \textbf{\begin{tabular}[c]{@{}c@{}}Image\\ Complexity\end{tabular}} & \textbf{Input}                                                                                                    & \textbf{\begin{tabular}[c]{@{}c@{}}Question\\ Type\end{tabular}} & \textbf{\begin{tabular}[c]{@{}c@{}}Task\\ Format\end{tabular}}               & \textbf{\begin{tabular}[c]{@{}c@{}}Question\\ Creation\end{tabular}} & \textbf{\begin{tabular}[c]{@{}c@{}}Segment\\ Creation\end{tabular}} \\ \hline \hline
\begin{tabular}[c]{@{}c@{}}Crossmodal-3600\\ \cite{thapliyal-etal-2022-crossmodal3600}\end{tabular}                     & 36               & -                 & 100              & 3,600          & -                 & Normal                                                              & \begin{tabular}[c]{@{}c@{}}Prompt +\\ An Image\end{tabular}                                                       & CU                                                               & \begin{tabular}[c]{@{}c@{}}Image\\ Captioning\end{tabular}                   & -                                                                    & -                                                            \\ \hline
\begin{tabular}[c]{@{}c@{}}MOSAIC\\ \cite{IPAS_mosaic_2025} \end{tabular}                               & -                & -                 & 336              & 1,500          & -                 & Normal                                                              & \begin{tabular}[c]{@{}c@{}}Prompt +\\ An Image\end{tabular}                                                       & CU                                                               & \begin{tabular}[c]{@{}c@{}}Image\\ Captioning\end{tabular}                   & -                                                                    & -                                                            \\ \hline
\begin{tabular}[c]{@{}c@{}}MosAIC \\ \cite{bai-etal-2025-power_MosAIC} \end{tabular}                             & 3                & 14                & 700              & 2,832          & -                 & Normal                                                              & \begin{tabular}[c]{@{}c@{}}Prompts +\\ An Image\end{tabular}                                                      & CU                                                               & \begin{tabular}[c]{@{}c@{}}Image\\ Captioning\end{tabular}                   & -                                                                    & -                                                            \\ \hline
\begin{tabular}[c]{@{}c@{}}SEA-VL\\ \cite{cahyawijaya2025crowdsourcecrawlgeneratecreating_SEA-VL} \end{tabular}                              & 11               & -                 & -                & 1.3M           & -                 & Normal                                                              & \begin{tabular}[c]{@{}c@{}}Prompts +\\ An Image\end{tabular}                                                      & CU                                                               & \begin{tabular}[c]{@{}c@{}}Image \\ Generation and\\ Captioning\end{tabular} & -                                                                    & -                                                            \\ \hline
\begin{tabular}[c]{@{}c@{}}MosAIG\\ \cite{bhalerao2025multiagentmultimodalmodelsmulticultural_mosaig} \end{tabular}                              & 5                & -                 & 25               & 9,000          & -                 & Normal                                                              & Prompt                                                                                                            & CU                                                               & \begin{tabular}[c]{@{}c@{}}Image \\ Generation\end{tabular}                  & -                                                                    & -                                                            \\ \hline \hline \hline
\begin{tabular}[c]{@{}c@{}}GD-VCR\\ \cite{yin2021broaden_GDVCR} \end{tabular}                              & 4                & -                 & 10               & 328            & 886               & Normal                                                              & \begin{tabular}[c]{@{}c@{}}Question + \\ An Image +\\ Textual Choices\end{tabular}                                & CU                                                               & MCVQA                                                                        & Human                                                                & -                                                            \\ \hline
\begin{tabular}[c]{@{}c@{}}MTVQA\\ \cite{tang2024mtvqa} \end{tabular}                               & 10               & 20                & -                & 2,116          & 6,778             & Normal                                                              & \begin{tabular}[c]{@{}c@{}}Question +\\ An Image\end{tabular}                                                     & CU                                                               & \begin{tabular}[c]{@{}c@{}}Open-ended \\ VQA\end{tabular}                    & Human                                                                & -                                                            \\ \hline
\begin{tabular}[c]{@{}c@{}}CVQA \\ \cite{mogrovejo2024cvqa} \end{tabular}                              & 30               & 10                & -                & 5,239          & 10,374            & Normal                                                              & \begin{tabular}[c]{@{}c@{}}Question + \\ An Image +\\ Textual Choices\end{tabular}                                & CU                                                               & MCVQA                                                                        & Human                                                                & -                                                            \\ \hline
\begin{tabular}[c]{@{}c@{}}CulturalVQA\\ \cite{nayak-etal-2024-benchmarking_CulturalVQA} \end{tabular}                         & 11               & 5                 & 13               & 2,328          & 2,328             & Normal                                                              & \begin{tabular}[c]{@{}c@{}}Question +\\ An Image\end{tabular}                                                     & CU                                                               & \begin{tabular}[c]{@{}c@{}}Open-ended \\ VQA\end{tabular}                    & \begin{tabular}[c]{@{}c@{}}AI +\\ Human\end{tabular}                 & -                                                            \\ \hline
\begin{tabular}[c]{@{}c@{}}CROPE \\ \cite{nikandrou_2025_crope} \end{tabular}                              & 5                & -                 & 158              & 1,060          & 1,060             & Normal                                                              & \begin{tabular}[c]{@{}c@{}}Question + \\ An Image +\\ Textual Choices\end{tabular}                                & CU                                                               & \begin{tabular}[c]{@{}c@{}}Binary\\ VQA\end{tabular}                         & Human                                                                & -                                                            \\ \hline
\begin{tabular}[c]{@{}c@{}}CVLUE-VQA \\ \cite{Wang_CVLUE_2025} \end{tabular}                            & 1                & 15                & 92               & 7,169          & 7,169             & Normal                                                              & \begin{tabular}[c]{@{}c@{}}Question +\\ An Image\end{tabular}                                                     & CU                                                               & \begin{tabular}[c]{@{}c@{}}Open-ended \\ VQA\end{tabular}                    & Human                                                                & -                                                            \\ \hline
\begin{tabular}[c]{@{}c@{}}CultureVerse-SR \&\\ CultureVerse-IR\\ \cite{liu2025culturevlm} \end{tabular}                     & 188              & 15                & 11,085           & 11,085         & 11,085            & Normal                                                              & \begin{tabular}[c]{@{}c@{}}Question + \\ An Image +\\ Textual Choices\end{tabular}                                & CU                                                               & MCVQA                                                                        & \begin{tabular}[c]{@{}c@{}}AI +\\ Human\end{tabular}                 & -                                                            \\ \hline
\begin{tabular}[c]{@{}c@{}}GIMMICK-COQA\\ \cite{schneider2025gimmickgloballyinclusive} \end{tabular}                        & 144              & 5                 & 728              & 6,857          & 982               & Normal                                                              & \begin{tabular}[c]{@{}c@{}}Question + \\ \# of Images +\\ Textual Choices\end{tabular}                            & CU                                                               & MCVQA                                                                        & \begin{tabular}[c]{@{}c@{}}AI +\\ Human\end{tabular}                 & -                                                            \\ \hline
\begin{tabular}[c]{@{}c@{}}MaRVL\\ \cite{liu-etal-2021-visually_MaRVL} \end{tabular}                               & 5                & 18                & 447              & 4,914          & 5,670             & Normal                                                              & \begin{tabular}[c]{@{}c@{}}Statement + \\ \# of Images +\\ Textual Choices\end{tabular}                           & CR                                                               & \begin{tabular}[c]{@{}c@{}}Binary\\ VQA\end{tabular}                         & Human                                                                & -                                                            \\ \hline
\begin{tabular}[c]{@{}c@{}}FoodieQA\\ \cite{li-etal-2024-foodieqa} \end{tabular}                            & 1                & 14                & -                & 389            & 403               & Normal                                                              & \begin{tabular}[c]{@{}c@{}}Question + \\ \# of Images as\\ Visual Choices\end{tabular}                            & CR                                                               & MCVQA                                                                        & Human                                                                & -                                                            \\ \hline
\begin{tabular}[c]{@{}c@{}}SEA-VQA\\ \cite{2024_seavqa} \end{tabular}                              & 8                & -                 & 53               & 515            & 1,999             & Normal                                                              & \begin{tabular}[c]{@{}c@{}}Question + \\ An Image +\\ Textual Choices\end{tabular}                                & CR                                                               & MCVQA                                                                        & \begin{tabular}[c]{@{}c@{}}AI +\\ Human\end{tabular}                 & -                                                            \\ \hline
\begin{tabular}[c]{@{}c@{}}K-Viscuit\\ \cite{corr_kviscuit} \end{tabular}                           & 1                & 10                & -                & 237            & 420               & Normal                                                              & \begin{tabular}[c]{@{}c@{}}Question + \\ An Image +\\ Textual Choices\end{tabular}                                & CR                                                               & MCVQA                                                                        & \begin{tabular}[c]{@{}c@{}}AI +\\ Human\end{tabular}                 & -                                                            \\ \hline
\begin{tabular}[c]{@{}c@{}}CultureVerse-CK\\ \cite{liu2025culturevlm} \end{tabular}                     & 188              & 15                & 11,085           & 11,085         & 11,085            & Normal                                                              & \begin{tabular}[c]{@{}c@{}}Question + \\ An Image +\\ Textual Choices\end{tabular}                                & CR                                                               & MCVQA                                                                        & \begin{tabular}[c]{@{}c@{}}AI +\\ Human\end{tabular}                 & -                                                            \\ \hline
\begin{tabular}[c]{@{}c@{}}GIMMICK-CIVQA\\ \cite{schneider2025gimmickgloballyinclusive} \end{tabular}                       & 144              & 5                 & 635              & 1,928          & 2,233             & Normal                                                              & \begin{tabular}[c]{@{}c@{}}Question + \\ An Image +\\ Textual Choices\end{tabular}                                & CR                                                               & MCVQA                                                                        & \begin{tabular}[c]{@{}c@{}}AI +\\ Human\end{tabular}                 & -                                                            \\ \hline
\begin{tabular}[c]{@{}c@{}}GIMMICK-CKQA\\ \cite{schneider2025gimmickgloballyinclusive} \end{tabular}                        & 144              & 5                 & 635              & 6,857          & 728               & Normal                                                              & \begin{tabular}[c]{@{}c@{}}Question + \\ An Image +\\ Textual Choices\end{tabular}                                & CR                                                               & MCVQA                                                                        & \begin{tabular}[c]{@{}c@{}}AI +\\ Human\end{tabular}                 & -                                                            \\ \hline \hline \hline
\begin{tabular}[c]{@{}c@{}}GlobalRG\\ \cite{bhatia-etal-2024-local_GlobalRG} \end{tabular}                            & 15               & 20                & 220              & 3,591          & -                 & Normal                                                              & \begin{tabular}[c]{@{}c@{}}Prompt +\\ An Image\end{tabular}                                                       & CU                                                               & \begin{tabular}[c]{@{}c@{}}Visual\\ Grounding\end{tabular}                   & -                                                                    & \begin{tabular}[c]{@{}c@{}}Human,\\ BBox\end{tabular}        \\ \hline
\begin{tabular}[c]{@{}c@{}}CVLUE-VG\\ \cite{Wang_CVLUE_2025} \end{tabular}                             & 1                & 15                & 92               & 7,169          & 5,385             & Normal                                                              & \begin{tabular}[c]{@{}c@{}}Prompt +\\ An Image\end{tabular}                                                       & CU                                                               & \begin{tabular}[c]{@{}c@{}}Visual\\ Grounding\end{tabular}                   & -                                                                    & \begin{tabular}[c]{@{}c@{}}Human,\\ BBox\end{tabular}        \\ \hline \hline \hline
\textbf{\begin{tabular}[c]{@{}c@{}}Seeing\\ Culture\\ Benchmark\\ (SCB)\end{tabular}} & 7                & 5                 & 138              & 1,065          & 3,178             & Complex                                                             & \begin{tabular}[c]{@{}c@{}}I) Question + \\ An Image +\\ Textual Choices\\ II) Question +\\ An Image\end{tabular} & CR                                                               & \begin{tabular}[c]{@{}c@{}}I) MCVQA,\\ II) Visual\\ Grounding\end{tabular}   & Human                                                                & \begin{tabular}[c]{@{}c@{}}Human,\\ Polygon\end{tabular}     \\ \hline
\end{tabular}%
}
\caption{Comparison between SCB and related works is divided into three distinct sections. The initial section addresses works that do not concentrate on VQA or visual grounding tasks. The subsequent portion focuses on VQA-related studies, while the final section pertains to visual grounding-related research. Here, "CU" stands for cultural understanding, and "CR" signifies cultural reasoning. "MCVQA" refers to multiple-choice VQA. We filter out images that depict only a single object or lack distractor objects, making our images complex compared to the others. This analysis underscores the distinctive contributions of SCB in furthering the development of cultural visual reasoning and grounding within the field.}
\label{tab:dataset_comparison}
\end{table*}

\section{Related Work}

\subsection{Benchmarks for Cultural Understanding}

The domain has seen the emergence of various recent multicultural vision-language datasets and benchmarks that incorporate explicit cultural taxonomies and tailored tasks (e.g., culture-aware VQA, grounding, and captioning), as shown in Table \ref{tab:dataset_comparison}. For example, Crossmodal-3600 \cite{thapliyal-etal-2022-crossmodal3600}, MOSAIC \cite{IPAS_mosaic_2025}, and MosAIC \cite{bai-etal-2025-power_MosAIC} are primarily centered on image captioning tasks. In contrast, while SEA-VL \cite{cahyawijaya2025crowdsourcecrawlgeneratecreating_SEA-VL} includes an image captioning component, its predominant emphasis is on image generation, similar to the approach taken by MosAIG \cite{bhalerao2025multiagentmultimodalmodelsmulticultural_mosaig}. 
Numerous studies examine VQA in various settings. For example, MTVQA \cite{tang2024mtvqa}, CulturalVQA \cite{nayak-etal-2024-benchmarking_CulturalVQA}, and a part of CVLUE \cite{Wang_CVLUE_2025} have open-ended questions, while CROPE \cite{nikandrou_2025_crope} employs binary (True/False) questions. More relevant to our work, GD-VCR \cite{yin2021broaden_GDVCR}, CVQA \cite{mogrovejo2024cvqa}, a part of CultureVerse \cite{liu2025culturevlm}, and a part of GIMMICK \cite{schneider2025gimmickgloballyinclusive} feature multiple-choice questions within the framework of cultural understanding. Unlike these studies that utilize textual options, our research incorporates visual alternatives. It is essential to note that we present SCB in a single row, whereas the results of some other studies are reported separately according to specific tasks. Our evaluation, however, combines two tasks, unlike the others, which evaluate each task separately. 
Besides, GlobalRG \cite{bhatia-etal-2024-local_GlobalRG} and a part of CVLUE \cite{Wang_CVLUE_2025} address the visual grounding of cultural artifacts using bounding boxes (BB), relying on straightforward prompts that include the keyword concept. In contrast, our research tackles questions that necessitate reasoning and employs a semantic segmentation mask that emphasizes fine-grained details. 

\subsection{Benchmarks for Cultural Reasoning}

Cultural reasoning is a critical aspect that distinguishes mere cultural understanding from deeper cognitive engagement with cultural contexts. From this perspective, various studies bridge the gap in the VQA task. For instance, MaRVL \cite{liu-etal-2021-visually_MaRVL} is the first dataset to focus on cultural reasoning; however, its objective is limited to determining the truth value of specific image captions.
SEA-VQA \cite{2024_seavqa}, K-Viscuit \cite{corr_kviscuit}, and a few parts of CultureVerse \cite{liu2025culturevlm} and GIMMICK \cite{schneider2025gimmickgloballyinclusive} focus on cultural reasoning through multiple-choice VQA. 
However, the multiple-choice responses in these studies are textual, and the questions are generated by AI, subsequently refined by human annotators, as seen in other related works. Additionally, unlike our study, these datasets lack a defined framework for selecting complex images, as discussed in Section \ref{create_image}.
Only FoodieQA \cite{li-etal-2024-foodieqa} offers visual options similar to our research and features human-constructed questions; however, it has a limited scope, focusing exclusively on Chinese cuisine. Moreover, the concept of visual grounding, which involves extracting evidence from an image to substantiate reasoning, has not been previously examined.

\section{SCB Benchmark}
\label{sec:scb}

Existing cultural benchmarks for Vision-Language Models (VLMs) exhibit several limitations, as detailed in Table \ref{tab:dataset_comparison}. 
In terms of these limitations, we observe the following: 1) the \textbf{questions} fail to foster both cultural reasoning and spatial grounding, 2) there is a scarcity of humanized \textbf{questions}, leading to a reliance on mechanical, AI-generated queries, 3) the \textbf{images} provided are often not sufficiently complex to challenge VLMs, e.g. lack of distractors. To address these challenges, the SCB provides a more nuanced approach by incorporating culturally rich images and authentic questions that reflect diverse cultural narratives. Further elaboration is provided in the respective sections. 

\begin{figure*}[!htb]
  \includegraphics[width=1\textwidth]{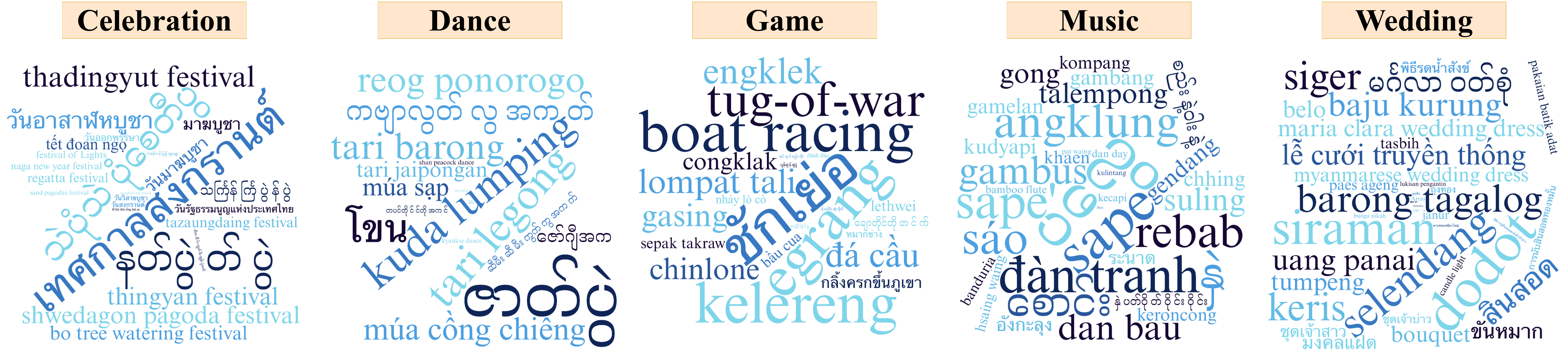}
  \caption{Word clouds illustrating the concepts of 1,093 unique questions in SCB are categorized into five cultural themes: wedding, game, music, celebration, and dance. The variation in font size within these clouds reflects the frequency of concept occurrences relevant to each theme. A simplified form for better visualization.}
  \label{fig:word_cloud}
\end{figure*}

\begin{figure*}[!htb]
  \includegraphics[width=0.33\textwidth]{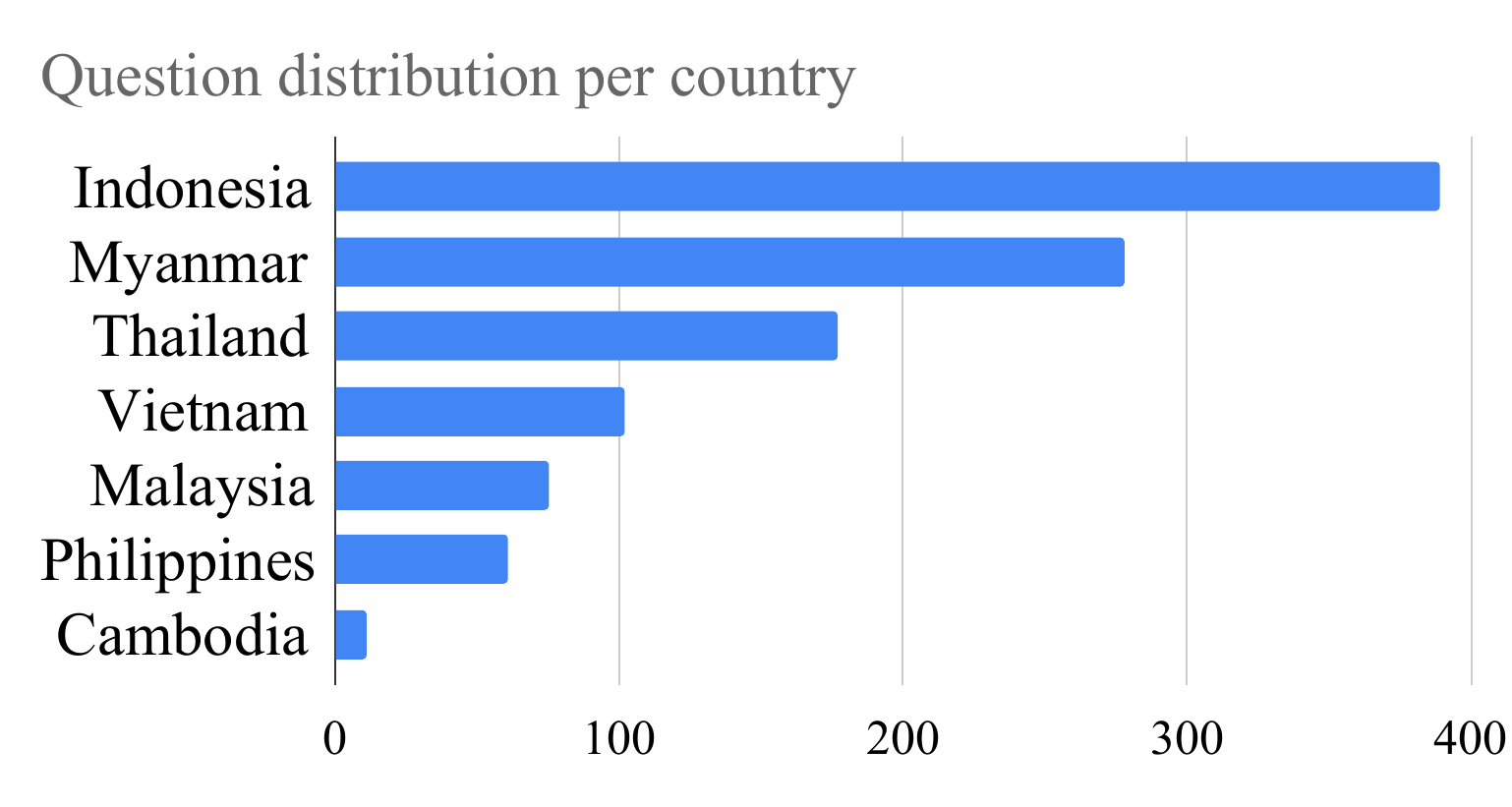}
  \includegraphics[width=0.33\textwidth]{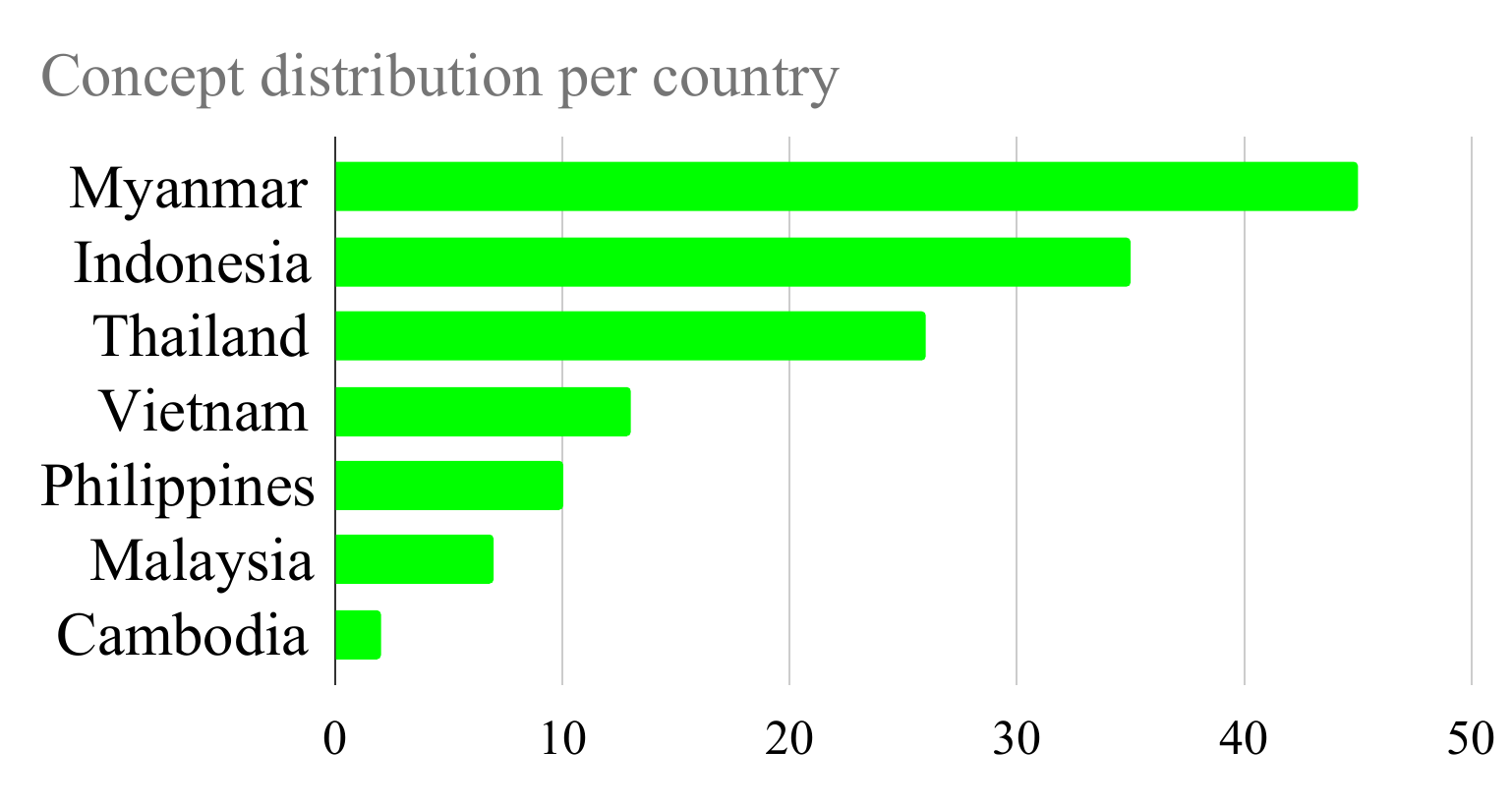}
  \includegraphics[width=0.33\textwidth]{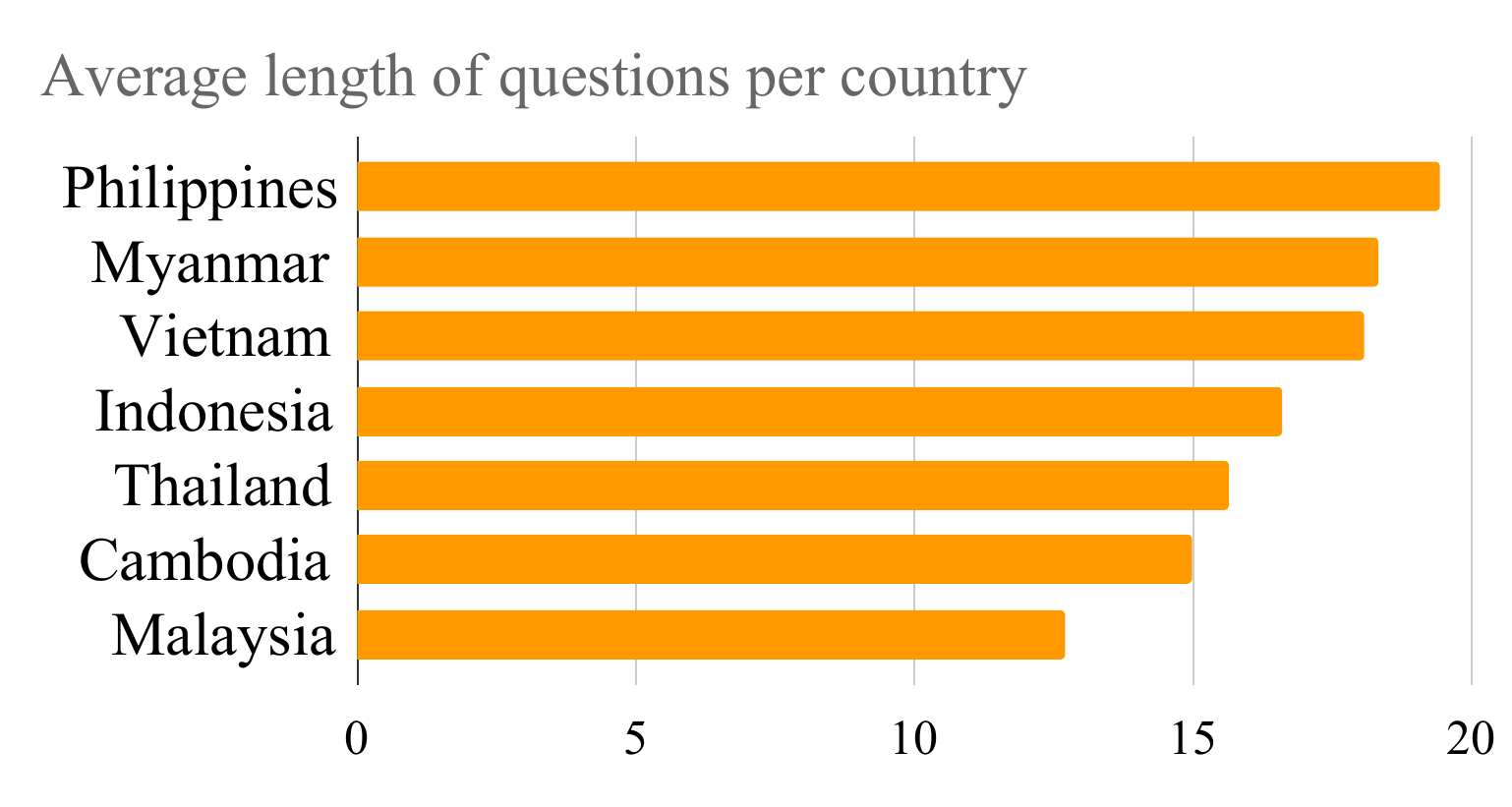}
  \includegraphics[width=0.33\textwidth]{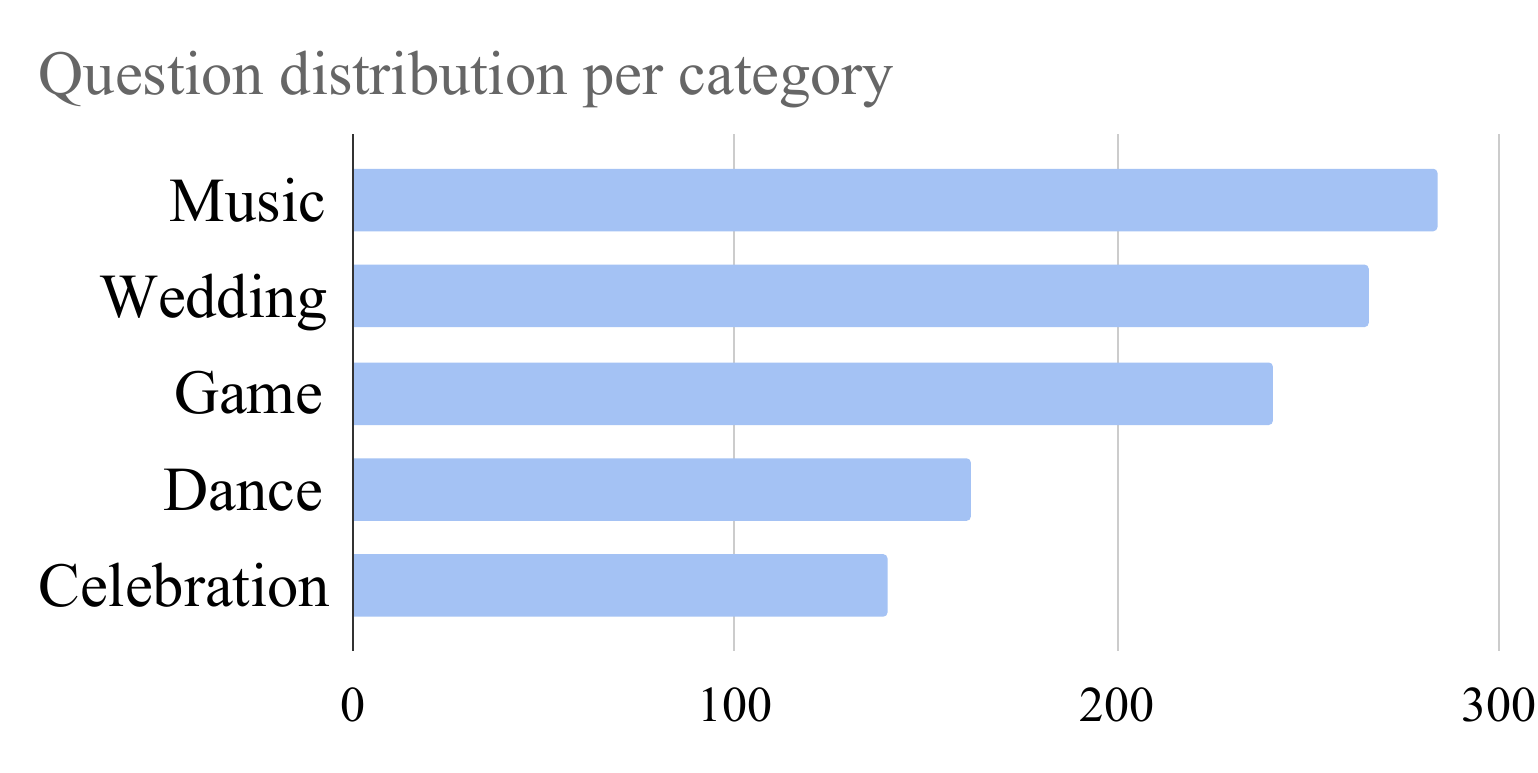}
  \includegraphics[width=0.33\textwidth]{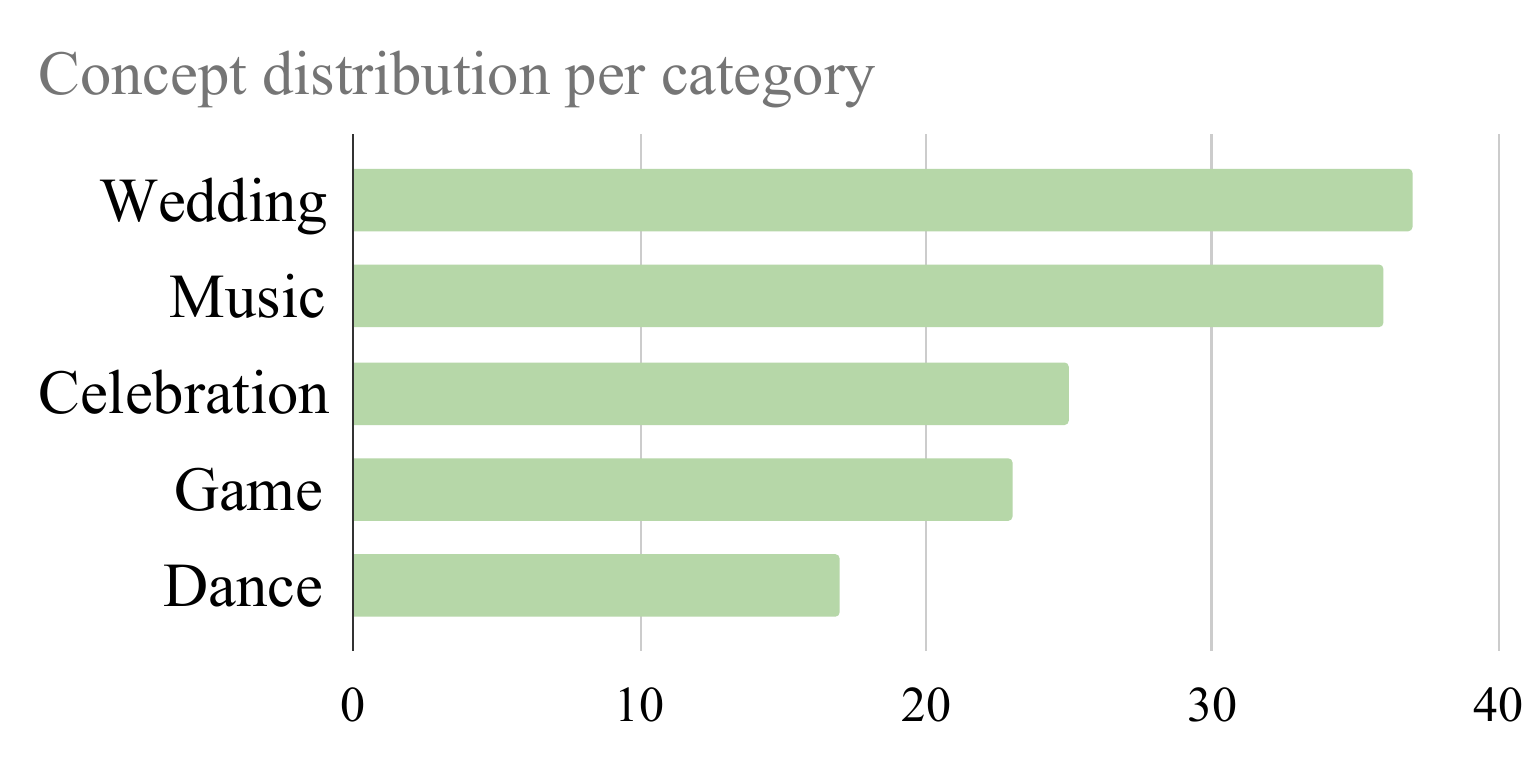}
  \includegraphics[width=0.33\textwidth]{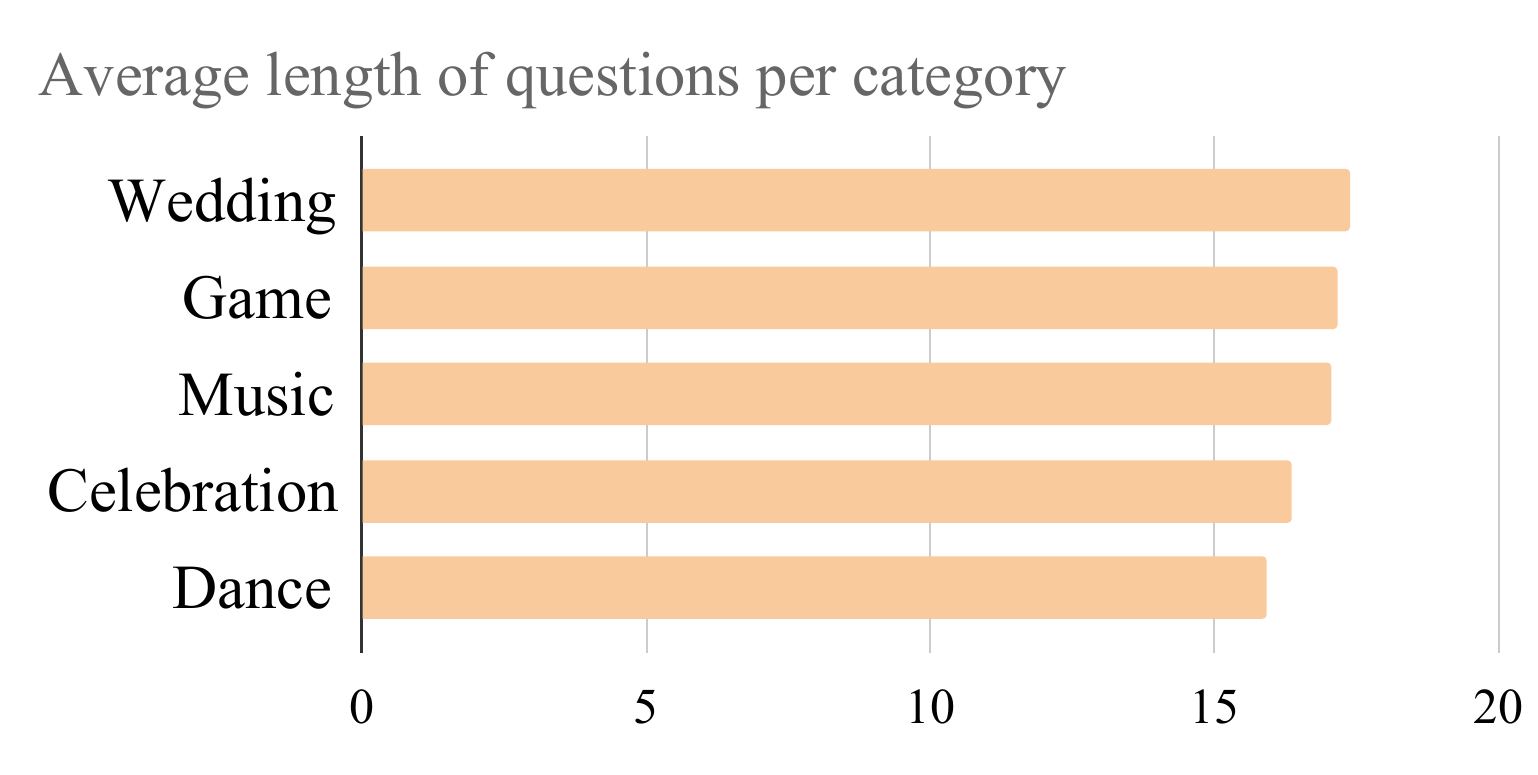}
  \caption{The figures encompass a comprehensive analysis of the distribution of unique questions, concepts, and the average length of questions, segmented by both country and category.}
  \label{fig:statistic_data}
\end{figure*}

\paragraph{Taxonomy.} We adopt a hierarchical framework to categorize cultural elements. Each national culture is subdivided into five principal categories: music, game, dance, celebration, and wedding. Within these categories, specific cultural concepts are delineated, allowing for a structured representation that can be expressed in the format of country/category/concept, \textit{e.g. Cambodia/music/khaen}. It is important to note that these categories are mutually exclusive; for instance, the \textit{music} category pertains solely to musical instruments, whereas the \textit{wedding} category encompasses garments and other cultural artifacts associated with the wedding ceremony. Additionally, some concepts may incorporate multiple characters or objects. For example, in Figure \ref{fig:intro}, the concept of the \textit{barong dance} includes two characters, \textit{barong} and \textit{monkey}. This approach facilitates a comprehensive understanding of cultural diversity and its manifestations across different societies. 

\paragraph{Countries.} To establish a benchmark that accurately encapsulates cultural diversity, we have selected seven underrepresented Southeast Asia countries, including Cambodia, Myanmar, Indonesia, Vietnam, the Philippines, Malaysia, and Thailand. This selection underscores the importance of recognizing and valuing the rich tapestry of cultural identities within this region. 

\paragraph{Concepts.} We solicit suggestions for cultural concepts based on the defined categories for each country using a Large Language Model (LLM), ChatGPT \cite{ChatGPT_software}. Following this, we conduct a survey to gather insights from local individuals representing each culture, either in English or their local language, to reach authentic images during the image crawling process. The survey aims to refine and validate the concepts proposed by the LLM, with two to three respondents from each country. Ultimately, we distill the results to identify concepts that receive unanimous agreement among the participants. A similar approach is applied to potential characters or objects associated with these concepts. A range of statistical visualizations regarding concepts and questions is presented in Figures \ref{fig:word_cloud} and \ref{fig:statistic_data}. 

\paragraph{Images.}
\label{create_image}
We crawl via Google Images based on the concepts we identify, collecting 150 images for each concept. Subsequently, we enlist human annotators to perform manual filtering to ensure the quality of the images. This filtration process assesses whether the retrieved images: i) are relevant to the concept keyword, ii) depict real-world scenarios, iii) are free from duplication, iv) do not have the cultural artifact completely or predominantly obscured, meaning images that are excessively focused on the cultural artifact with a blurry background are excluded, v) contain various distracting objects or scenes, preferably related other cultural artifacts, which may cause conflict to other cultural concept(s) vi) yet sufficiently clear to identify the cultural artifact. The initial three steps, which are standard practice in other datasets, reduce the image count from approximately 20,000 to 4,000. Nonetheless, the final three steps distinguish our image-collecting process. We also incorporate 32 images from the SEA-VL \cite{cahyawijaya2025crowdsourcecrawlgeneratecreating_SEA-VL} dataset. Ultimately, through a meticulous review, we ensure that the SCB comprises 1,065 unique images.

\paragraph{Segmentation.} Upon selecting the images, annotators use an online segmentation tool \cite{make-sense-ai} to segment the corresponding concept keywords or their associated cultural artifacts, such as characters in a local dance or objects used for specific celebrations. This can be illustrated in Figure \ref{fig:sample}, particularly in the segments denoted as \textit{Indonesia/dance/balinese legong} and \textit{Thailand/celebration/songkran festival}. Note that segmentation is performed using polygons instead of bounding boxes to ensure the capture of intricate details. 


\paragraph{Question Formulation.} We instruct annotators to formulate unique questions that are culturally aligned with the specific artifacts segmented in the images, while refraining from using templates. Specifically, questions should not refer directly to the artifact itself but rather to the symbols or cultural significance associated with it. Annotators are instructed to rely solely on their cultural knowledge, deliberately excluding any AI-generated sources. This ensures that each question requires a deeper reasoning of the culture authentically. For instance, the question, "In a traditional Thai wedding, what symbolizes the spiritual connection and blessings given to the couple by elders or religious figures?", pertains to the artifact represented by \textit{Thailand/wedding/double auspicious headband}, which is accompanied by a prompt of "Locate the artifact in the image." as well. Subsequently, annotators adapt the questions into a VQA format. Following the same line of questioning, this can be rephrased as: "Which image is associated with a traditional Thai wedding artifact that symbolizes the spiritual connection and blessings given to the couple by elders or religious figures?" This is further refined by omitting the segmentation-oriented prompt. In addition, annotators are tasked with providing a rationale for the correct answer, drawing from either online resources or their own cultural knowledge.    

\paragraph{Multiple-Choice Questions and Visual Options.} We extend these unique multiple-choice VQA questions into three types, utilizing varying visual options in our selection process. The basis of this approach is to utilize the same question paired with its corresponding correct answer. In contrast, the incorrect options are selected using three distinct pooling strategies: \underline{Type 1 (within culture)}, which sample concepts within the same category and country, \underline{Type 2 (across culture)}, which sample concepts within same category but completely different country for all options, and \underline{Type 3 (mix culture)}, which consists of balanced mix of Type 1 and Type 2 through a rule-based choice-swapping. For instance, for each randomly chosen pair of options from the Type 1 question, including the ground truth (GT) choice, we randomly sample the other two options from Type 2 questions, ensuring a balanced representation of options. To mitigate potential biases in this combination, each question is limited to a maximum of two repetitions for Type 3.
The number of images for visual options is capped at 20 for all types. See Appendix \ref{subsec:algorithm} for the algorithms.


\begin{table*}[!htb]
\centering
\resizebox{0.95\textwidth}{!}{%
\begin{tabular}{lcccccccc}
\toprule
Model & \multicolumn{2}{c}{Type 1} & \multicolumn{2}{c}{Type 2} & \multicolumn{2}{c}{Type 3} & \multicolumn{2}{c}{Overall} \\
\cmidrule(lr){2-3} \cmidrule(lr){4-5} \cmidrule(lr){6-7} \cmidrule(lr){8-9}
 & Acc & Mean IoU & Acc & Mean IoU & Acc & Mean IoU & Acc & Mean IoU \\
\midrule
InstructBLIP       & 11.07 & --   & 10.31 & --   & 11.04 & --   & 10.86 & --   \\
Idefics2           & 13.21 & 0.19 & 11.03 & 0.05 & 12.30 & 0.18 & 12.21 & 0.15 \\
Llama-3.2          & 23.57 & --   & 25.66 & --   & 23.80 & --   & 24.23 & --   \\
LLaVA-Onevision    & 26.43 & --   & 25.18 & --   & 23.47 & --   & 24.70 & --   \\
MiniCPM-2.6        & 28.33 & --   & 34.65 & --   & 32.85 & --   & 32.13 & --   \\
InternVL2.5-4B     & 30.83 & 28.37& 30.34 & 28.88& 32.18 & 28.49& 31.34 & 28.56\\
Qwen2.5-VL-7B      & 44.17 & \textbf{44.90}& 61.51 & \textbf{48.22}& 54.85 & \textbf{47.60}& 53.78 & \textbf{47.20}\\ \midrule
GPT-4.1            & 68.33 & 13.31& 90.17 & 14.32& 85.04 & 13.60& 81.97 & 13.74\\
Gemini-2.5-Pro     & 71.07 & 16.56& 90.17 & 16.67& 85.44 & 15.79& 82.88 & 16.22\\
GPT-o3            & \textbf{73.69} & 31.10& \textbf{91.13} & 32.50& \textbf{88.23} & 31.69& \textbf{85.15} & 31.78\\
\bottomrule
\end{tabular}
}
\caption{Detailed performance benchmark with several VLMs on our Visual Reasoning and Grounding task. The upper section focuses on open-source VLMs, whereas the lower section pertains to closed-source models. Type 1 is defined as \textit{within culture}, Type 2 as \textit{across culture}, and Type 3 represents a balanced combination of both Type 1 and Type 2.}
\label{tab:perf}
\end{table*}

\begin{figure*}[!htb]
  \centering
  \includegraphics[width=0.95\textwidth]{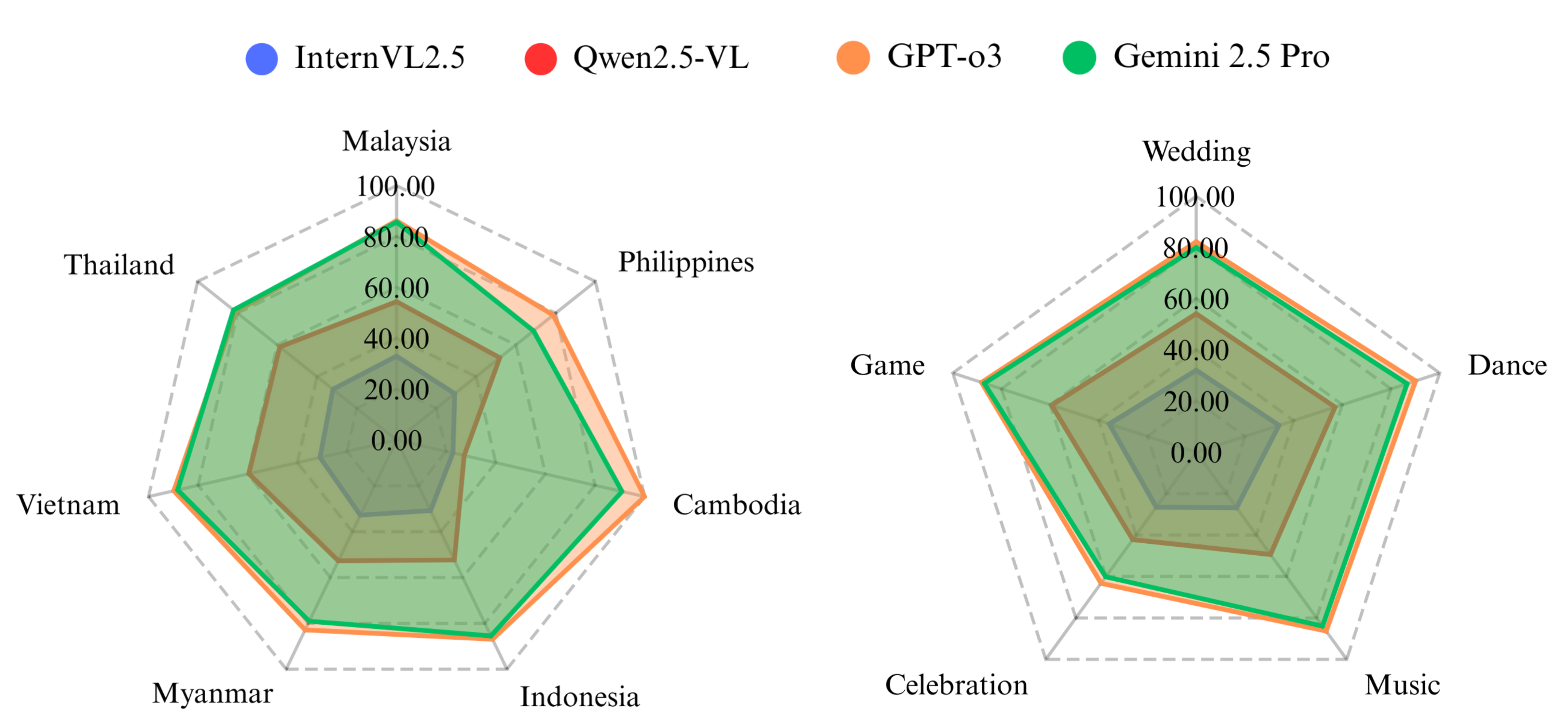}
  \caption{The overall multiple-choice VQA accuracy of certain VLMs across different countries and categories.}
  \label{fig:res_radar}
\end{figure*}

\section{Experiments}

\subsection{Visual Reasoning and Grounding Task}


We perform a zero-shot evaluation utilizing the following prompt in the initial phase: a textual question for VQA alongside visual options. The output corresponds to one of the provided image options. To assess performance, we employ accuracy as the metric, in accordance with established methodologies in multiple-choice VQA tasks \cite{7780909, nayak-etal-2024-benchmarking_CulturalVQA}. In the initial phase, questions that are accurately addressed with the appropriate visual option advance to a subsequent stage to segment the cultural artifacts, while those that are not are excluded.
In the following phase, given an image $I$ and a question $q$ that pertains to a cultural term, the objective is to generate a segmentation mask $R$ that delineates the area in $I$ relevant to $q$. We evaluate performance using bounding boxes (BB) rather than polygons, as current VLMs capable of both VQA and segmentation are restricted to grounding at the BB level. Consequently, the performance of the models is assessed by measuring the overlap between the predicted regions of interest and GT masks, employing Intersection over Union (IoU) as the evaluation metric: $IoU = \frac{R\cap R_{GT}}{R\cup R_{GT}}$. We then report it as the mean IoU. 

\begin{figure*}[!htb]
  \includegraphics[width=1\textwidth]{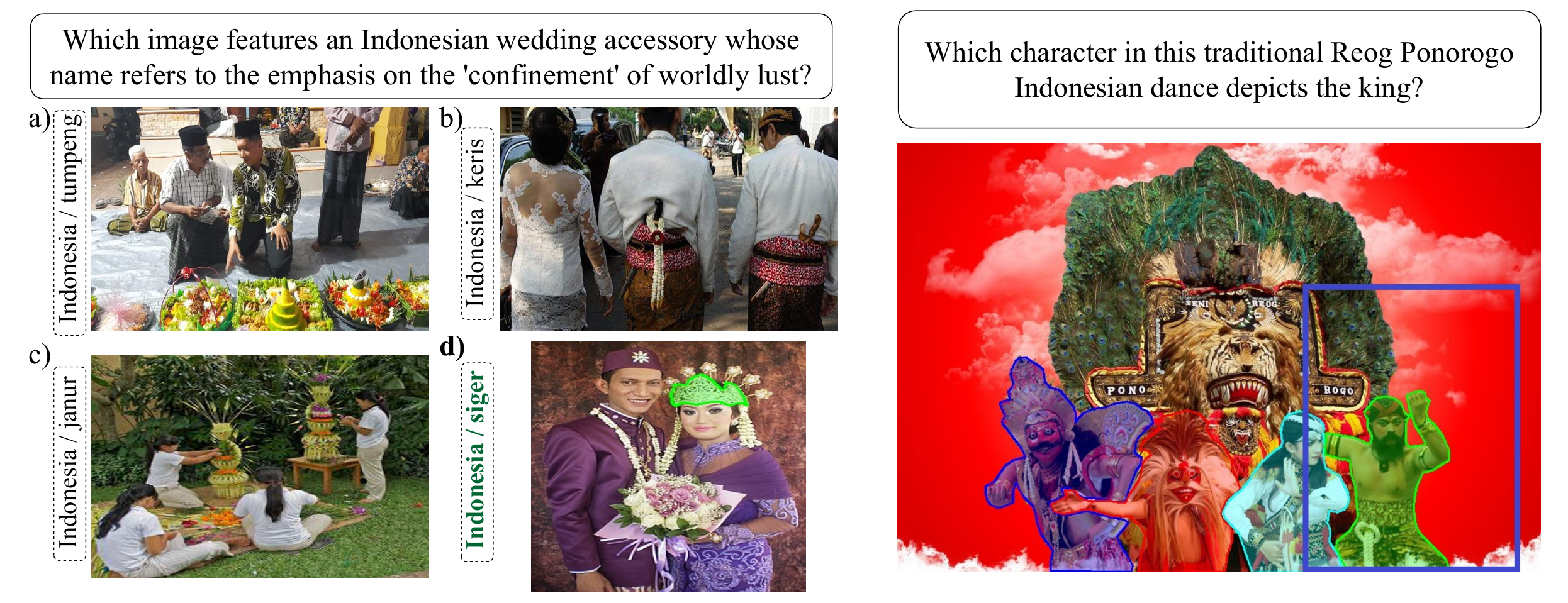}
  \caption{The figure presents two examples of failures for each stage. The left side illustrates an example of multiple-choice VQA, where all VLMs fail to select the correct option. Conversely, the right side pertains to the spatial grounding, for another example. Notably, this specific output is generated by GPT-o3, which is the only VLM that accurately answers the multiple-choice VQA version of this spatial grounding question. The blue character on the far left identifies the correct segment, while GPT-o3 incorrectly selects the option on the far right.}
  \label{fig:qualitative}
\end{figure*}

\paragraph{Implementation details.} 
We conduct a comparative analysis of various advanced VLMs. This includes closed-source models such as GPT-4.1, GPT-o3 and Gemini Pro 2.5, alongside a diverse selection of open-source models that vary in size: IntructBLIP 7B \cite{dai2023instructblip}, Idefics2 8B \cite{idefix2}, LLama 3.2 11B \cite{llama32}, LLaVa-OneVision 7B \cite{li2025llavaonevision}, MiniCPM 2.6 8B \cite{yao2024minicpm}, InternVL2.5 4B \cite{chen2024expanding} and Qwen2.5-VL 7B \cite{qwen25}. It is essential to note that we do not employ VLMs capable of segmentation but not suited for multiple-choice VQA, given the requirements of our task. 
The quantity of Type 1, Type 2, and Type 3 questions is 834, 840, and 1,504, respectively. Our analysis is constrained by the operational parameters of our multiple-choice VQA generation algorithms. Considering that an image cannot be selected as an answer option for more than a certain number in each type, some questions do not have enough answer options, and we omit those questions. In this regard, 259 questions from Type 1 and 253 questions from Type 2 are excluded from our analysis due to this criterion, given the unique total of 1,093 questions. 
The higher number of Type 3 questions results from our allowance for repeating questions up to a maximum of two times, in line with the aforementioned algorithm. We first identified 871 unique Type 3 questions. Following the implementation of repetitions, we generated an additional 633 questions, adhering to the established constraints, which culminated in a total of 1,504 Type 3 questions.

\subsection{Results} 

\paragraph{How do VLMs' performance vary across different question types?} The findings presented in Table \ref{tab:perf} reveal that VLMs, both open-source and closed-source, exhibit their poorest performance when the visual options originate from the same country, whereas they display the highest performance when the visual options come from different countries. This pattern can largely be explained by the contextual clues embedded in the questions that pertain to specific countries or cultures. As a result, VLMs are more adept at eliminating alternative visual options that may include indicators from diverse countries. Notably, the correct answer choices (a, b, c, and d) are evenly distributed in our multiple-choice VQA dataset, each accounting for approximately 24\% to 26\% of the total. This distribution remains consistent across all subsets. Based on this distribution, the expected accuracy of random guessing is approximately 25\%. Furthermore, it is observed that 8.5\% of the multiple-choice questions are consistently answered incorrectly by all three closed-source models.

\paragraph{Can VLMs validate their reasoning by segmenting the cultural artifact?} A notable discrepancy exists between visual reasoning capabilities and spatial grounding. For example, while GPT-o3 achieves an accuracy exceeding 90\%, its mIoU score does not surpass 33\%. This disparity is even more pronounced in other closed-source VLMs. Conversely, Qwen exhibits a smaller gap, considering its superior spatial grounding performance and lower efficacy in multiple-choice VQA. Overall, this suggests that, although VLMs may frequently select the correct answer, they often fail to ground their reasoning adequately.
We further investigate whether this phenomenon suggests that VLMs possess limited object segmentation capabilities. We perform grounding by referring to cultural objects instead of reasoning for Type 1, and the mean IoU is as follows: Qwen 62.46, Gemini 30.80, GPT-o3 46.98. Compared to Table \ref{tab:perf}, grounding by reasoning results in an average drop of 16\% in the mean IoU. The result shows that this phenomenon is not solely due to VLMs’ segmentation skill. Moreover, a recent work \cite{Wang_CVLUE_2025} also shows that VLMs can reach 80\% and 40\% accuracy in segmenting general and cultural objects, respectively.

\paragraph{Do VLMs perform better in specific countries and categories?} As illustrated in Figure \ref{fig:res_radar} regarding the multiple-choice VQA stage, Qwen demonstrates superior performance when compared to other open-source VLMs; however, it still significantly trails behind GPT-o3. Notably, GPT-o3 achieves its highest performance in Cambodia, whereas Qwen performs least effectively in the same country. The remaining open-source models are considerably less performant than Qwen and display relatively varied outcomes among themselves. Besides, VLMs generally perform the best at \textit{dance} while performing the worst in \textit{celebration}. The dance category primarily features specific dancer characters, while the celebration category encompasses cultural artifacts that represent intangible concepts.

\paragraph{Qualitative results.} Figure \ref{fig:qualitative} presents examples of failures. The left side image illustrates that all presented VLMs are unable to select the appropriate visual option within the same country. The prediction is easier for options involving multiple objects, as seen on the right side, due to more distinguishable image features. In contrast, visual grounding is more difficult because similar yet distinct candidates can confuse the model. Specifically, GPT-o3 correctly selects the correct option but fails to identify the supporting evidence (blue mask), instead predicting the location of `Warok' (blue box). Overall, GPT-o3 achieves an MCQ accuracy of 94.79\% on this query type—higher than its performance on all other query types—while its mIoU is 27.33\%, the lowest among all.
More results and details can be found in the Appendix, such as Table \ref{tab:result_country} and \ref{tab:result_category}.

\paragraph{Further results with the equal distributions of Type 1, Type 2, and Type 3 questions.}

Due to the nature of their design, the number of questions in these three types is different. We further analyze the results when the sample sizes of the three question types are equal. This is achieved by selecting 664 identical samples from each type. The performances are similar to the results reported in Table 2, leading to a consistent conclusion.

\begin{table}[!htb]
\centering
\resizebox{\columnwidth}{!}{%
\begin{tabular}{lllllll}
\hline
\multirow{2}{*}{Model} & \multicolumn{2}{c}{Type 1}      & \multicolumn{2}{c}{Type 2}      & \multicolumn{2}{c}{Type 3}      \\ \cline{2-7} 
 & \multicolumn{1}{c}{Acc} & \multicolumn{1}{c}{mIoU} & \multicolumn{1}{c}{Acc} & \multicolumn{1}{c}{mIoU} & \multicolumn{1}{c}{Acc} & \multicolumn{1}{c}{mIoU} \\ \hline
Qwen          & 44.58          & \textbf{47.75} & 60.69          & \textbf{49.38} & 53.78          & \textbf{48.22} \\ \hline
Gemini         & 73.34          & 16.32          & 90.51          & 16.45          & 85.49          & 15.72          \\
GPT-o3                 & \textbf{75.75} & 32.40          & \textbf{92.17} & 32.77          & \textbf{88.43} & 31.35          \\ \hline
\end{tabular}%
}
\caption{Additional findings utilizing the same sample sets from each question category in our Visual Reasoning and Grounding task are presented. The upper segment concentrates on open-source VLMs, while the lower segment addresses closed-source models. \textit{Qwen} denotes Qwen2.5-VL-7B, and \textit{Gemini} refers to Gemini-2.5-Pro.}
\label{tab:perf-equa}
\end{table}

\section{Conclusion}

In conclusion, this paper presents the Seeing Culture Benchmark (SCB), which addresses the need for improved cultural reasoning in multimodal VLMs. By employing a two-stage approach that incorporates VQA and cultural artifact segmentation, we provide a framework for assessing VLMs on culturally rich images from seven Southeast Asia countries. Our dataset includes 1,065 images and 3,178 curated questions, highlighting the underrepresented cultural diversity of the region. Our findings reveal the significant challenges of cross-modal cultural reasoning, emphasizing the need for enhanced visual reasoning and spatial grounding in culturally nuanced contexts. SCB is a vital resource for advancing research in this domain and addressing identified shortcomings in existing VLMs.

\section*{Acknowledgement} 
This research is supported by the Ministry of Education, Singapore, under its Academic Research Fund Tier 2 (Proposal ID: T2EP20222-0047). Any opinions, findings, and conclusions or recommendations expressed in this material are those of the authors and do not reflect the views of the Ministry of Education, Singapore. Professor NGO Chong Wah gratefully acknowledges the support of the Lee Kong Chian Professorship awarded by Singapore Management University.

\section*{Limitation} 

We acknowledge several constraints in our approach as outlined below.

\paragraph{Cultural Representation.} Our objective was to encompass all countries in Southeast Asia; however, we faced challenges in sourcing sufficient cultural concepts through data crawling and in locating adequately qualified human annotators who align with the requirements outlined in the paper from specific nations, including Timor-Leste, Brunei, and Laos. 

\paragraph{Long-tailed Distribution.} The aforementioned issues related to the availability of qualified human annotators from certain regions who align with the requirements outlined in our paper have proven challenging.
Furthermore, difficulties in acquiring high-quality images that fulfill our stringent filtration criteria for specific categories and countries, such as Cambodia, have resulted in a naturally occurring long-tailed distribution. 

\paragraph{Scalability.} This study entirely relies on human-generated questions, which are not suitable for scaling. However, we also consider a semi-automated approach for future work, which uses human-crafted questions as seeds to scale up our dataset in terms of the number of images, questions, and cultural representations. This could also mitigate the aforementioned limitations.

\section*{Ethical Consideration}

\paragraph{Cultural concepts overlap across cultures.} Certain cultural artifacts are commonly found in multiple countries, albeit with nuanced differences, characterized by the use of either identical or distinct cultural concept terminology. To mitigate potential conflicts, we implemented an "avoid list" during the selection of visual options for the question types. This initial measure effectively reduced the total number of questions from over 1,000 to more than 800 for both Type 1 and Type 2 questions; however, it also contributed to the overall stability of our research framework.

\paragraph{Annotators.} We recruited annotators through Upwork, a global freelancing platform, following specific criteria. Firstly, participants were required to be natives of Southeast Asian countries, possessing a comprehensive understanding of the local culture, traditions, and customs. Secondly, they needed to have a basic proficiency in using computers or mobile devices, as they were expected to utilize specialized software for image labeling. We employed purposive sampling to identify freelancers on Upwork.com who fulfilled these inclusion criteria, focusing on their cultural expertise and experience with cultural content or research. 

Additionally, potential participants were evaluated based on their profiles, work history, reviews, and portfolio samples, with a priority given to those who demonstrated a strong grasp of local culture and relevant project experience. This methodology ensures that selected participants not only possess knowledge of their cultural background but also have the necessary skills to utilize the required tools and adhere to the research protocols. For our study, we engaged three annotators each for the Philippines and Myanmar, and two annotators for the remaining countries. Participants were compensated monetarily at a rate of \$5-10 per hour for their involvement in the research, with specific compensation structured at \$5 for every 50 images labeled accurately.

\paragraph{Privacy Rights.} 
We ensure that the intellectual property and privacy rights of the images collected are respected. We claim that the collected data will not be used commercially. Our process involves retrieving images through Google Image Search, leading us to a variety of publicly available sources, including news websites, academic repositories, Wikipedia, and cultural heritage sites such as Wikimedia Commons and various encyclopedias. Although we do not employ specific filtering mechanisms for image licensing, we diligently retain and disclose all source URLs to guarantee complete traceability and transparency regarding image origins. Hence, we release the dataset under the CC BY-NC-SA 4.0 license, making the questions, image annotations, and license-free images publicly accessible through HuggingFace. The license stipulates that its use is restricted to non-commercial research purposes, allowing for deployment only with appropriate attribution and adherence to share-alike principles. This framework facilitates responsible downstream utilization while respecting the rights of the sources through explicit citation requirements embedded in the dataset's metadata. For copyrighted images, we share the source URLs. Also, we ensure that the dataset contains no personally identifiable information. The questions only cover cultural concepts, and the image annotations contain only the polygons of cultural artifacts.

\bibliography{main}

\appendix

\section{Appendix}
\label{sec:appendix}

\subsection{More Quantitative Results}
\label{subsec:more-results}

Table \ref{tab:result_country} and Table \ref{tab:result_category} display the full details for the overall results for \textit{country} and \textit{category}. We observe that closed-source VLMs generally exhibit higher accuracy performance, while open-source ones achieve higher mIoU results.

\begin{table*}[!htb]
\centering
\resizebox{1\textwidth}{!}{%
\begin{tabular}{l *{7}{cc}}
\toprule
Model
  & \multicolumn{2}{c}{Malaysia}
  & \multicolumn{2}{c}{Philippines}
  & \multicolumn{2}{c}{Cambodia}
  & \multicolumn{2}{c}{Indonesia}
  & \multicolumn{2}{c}{Myanmar}
  & \multicolumn{2}{c}{Vietnam}
  & \multicolumn{2}{c}{Thailand} \\
\cmidrule(lr){2-3}
\cmidrule(lr){4-5}
\cmidrule(lr){6-7}
\cmidrule(lr){8-9}
\cmidrule(lr){10-11}
\cmidrule(lr){12-13}
\cmidrule(lr){14-15}
  & Acc   & mIoU    & Acc    & mIoU   & Acc    & mIoU    & Acc    & mIoU    & Acc    & mIoU    & Acc    & mIoU    & Acc    & mIoU    \\
\midrule
InstructBLIP    &  3.85 & --     &  8.96 & --     &  4.55 & --     & 10.6  & --     & 12.97 & --     & 10.19 & --     & 13.37 & --     \\
Idefics2        &  9.89 & 0.03   &  2.83 & --     &  9.09 & --   & 12.42 & 0.22   & 15.20 & 0.11   & 14.51 & 0.19   & 10.17 & 0      \\
Llama 3.2       & 26.37 & --     & 21.23 & --     & 13.64 & --     & 24.40 & --     & 22.73 & --     & 25.93 & --     & 26.45 & --     \\
LLaVA-Onevision & 30.22 & --     & 28.77 & --     & 13.64 & --     & 23.89 & --     & 23.15 & --     & 23.46 & --     & 27.62 & --     \\
MiniCPM 2.6     & 44.51 & --     & 22.17 & --     & 18.18 & --     & 35.08 & --     & 25.38 & --     & 28.09 & --     & 38.66 & --     \\
InternVL2.5     & 32.97 & 33.35  & 29.25 & 32.47  & 22.73 & 22.27  & 30.79 & 29.42  & 32.78 & 26.61  & 30.86 & 32.19  & 31.98 & 21.57  \\
Qwen2.5-VL      & 54.40 & \textbf{56.89}  & 51.89 & \textbf{52.78}  & 27.27 & \textbf{60.50}  & 52.36 & \textbf{46.47}  & 52.72 & \textbf{48.55}  & 59.57 & \textbf{45.56}  & 58.72 & \textbf{40.65}  \\ \hline
GPT-4.1         & 84.07 & 14.64  & 69.81 & 16.79 & \textbf{100.00} & 18.19  & 85.19 & 13.82  & 77.27 & 16.64  & 86.73 &  7.33  & 79.65 & 11.60  \\
Gemini 2.5 Pro  & 85.71 & 17.10  & 68.87 & 15.40  & 90.91 & 13.36  & 85.48 & 15.76  & 79.08 & 20.21  & 88.27 & 12.53  & \textbf{81.98} & 13.98  \\
GPT-o3          & \textbf{86.26} & 41.74  & \textbf{78.77} & 35.29  & \textbf{100.00} & 30.23 & \textbf{86.93} & 32.77  & \textbf{82.85} & 31.51  & \textbf{89.81} & 28.13  & 80.81 & 24.31  \\
\bottomrule
\end{tabular}
}
\caption{The comprehensive performance of vision-language models (VLMs) is depicted by country. The term "Acc" signifies \textit{accuracy}, whereas "mIoU" stands for \textit{mean Intersection over Union}. Values in bold represent the highest figures within their respective columns.}
\label{tab:result_country}
\end{table*}

\begin{table*}[!htb]
\centering
\resizebox{1.7\columnwidth}{!}{%
\begin{tabular}{l *{5}{cc}}
\toprule
Model
  & \multicolumn{2}{c}{Wedding}
  & \multicolumn{2}{c}{Dance}
  & \multicolumn{2}{c}{Music}
  & \multicolumn{2}{c}{Celebration}
  & \multicolumn{2}{c}{Game} \\
\cmidrule(lr){2-3}
\cmidrule(lr){4-5}
\cmidrule(lr){6-7}
\cmidrule(lr){8-9}
\cmidrule(lr){10-11}
  & Acc    & mIoU   & Acc    & mIoU   & Acc    & mIoU   & Acc    & mIoU   & Acc    & mIoU   \\
\midrule
InstructBLIP    & 12.09 & --     & 18.42 & --     &  5.03 & --     & 18.75 & --     & 10.76 & --     \\
Idefics2        &  7.85 & 0.14   & 16.01 & 0.06   & 12.37 & --     & 17.19 & 0.25   & 14.49 & 0.22   \\
Llama 3.2       & 25.45 & --     & 25.44 & --     & 22.01 & --     & 32.03 & --     & 23.39 & --     \\
LLaVA           & 26.19 & --     & 24.78 & --     & 23.79 & --     & 24.22 & --     & 23.96 & --     \\
MiniCPM         & 33.40 & --     & 36.40 & --     & 35.43 & --     & 21.88 & --     & 24.96 & --     \\
InternVL        & 32.03 & 25.47  & 33.99 & 37.06  & 26.83 & 36.48  & 26.56 & 15.36  & 35.72 & 20.68  \\
Qwen            & 54.08 & \textbf{40.91}  & 57.02 & \textbf{48.97}  & 49.37 & \textbf{54.43}  & 42.19 & \textbf{31.63}  & 59.40 & \textbf{47.63}  \\ \hline
GPT-4.1         & 81.12 & 13.29  & 85.53 & 15.54  & 81.76 & 14.53  & 62.50 & 15.36  & 84.65 & 11.87  \\
Gemini          & 79.96 & 13.72  & 86.62 & 20.17  & 83.96 & 19.30  & 60.16 & 18.26  & 87.09 & 12.41  \\
GPT-o3          & \textbf{82.18} & 28.33  & \textbf{90.13} & 39.51  & \textbf{86.37} & 37.06  & \textbf{63.28} & 21.68  & \textbf{88.24} & 25.23  \\
\bottomrule
\end{tabular}
}
\caption{The overall effectiveness of vision-language models (VLMs) is illustrated across various categories. The abbreviation "Acc" denotes \textit{accuracy}, while "mIoU" refers to \textit{mean Intersection over Union}. Bolded values indicate the highest results within each respective column.}
\label{tab:result_category}
\end{table*}

\subsection{Seeing Culture Benchmark}
\label{subsec:dataset-appendix}

\subsubsection{Concepts} Figure \ref{fig:all_list} presents all the concepts, while Figure \ref{fig:mcq_example} shares more examples from our SCB dataset.

\subsubsection{Eliminated images and questions}

In accordance with the details outlined in Section \ref{sec:scb}, we exclude certain images from consideration. Specifically, as shown in Figure \ref{fig:eliminated_img}, we remove the image on the left as its focus is solely on the target cultural artifact. The image on the right is also omitted due to the lack of a distracting object, although it contains a more complex scene than the image on the left.
Besides, specific questions are excluded due to their generic nature, potential overlap with other cultural artifacts, or lack of necessity for critical reasoning. For example, we dismissed the question concerning \textit{Indonesia/game/permainan kelereng}: "Which object in the image symbolizes childhood nostalgia, often played in schoolyards and neighborhoods in Indonesia?" because numerous games evoke similar childhood memories. 
Similarly, we rejected the question for \textit{Myanmar/music/myanmarese saung}: "Which Burmese object in the image has a hollow body made of wood, designed to enhance the richness of its sound?" as it merely describes the cultural artifact without engaging in reasoning or referencing a symbol.

\subsubsection{Multiple-choice VQA Generation Algorithm}
\label{subsec:algorithm}

Algorithms \ref{algo:1}, \ref{algo:2}, and \ref{algo:3} explain how we choose visual options for each type. Additionally, we provide clarifications for the abbreviations utilized within the algorithms.

\begin{itemize}
  \item $\mathcal{D}$: Dataset
  \item $\mathcal{V}$: Vectorstore index
  \item $k$: Number of similar items to retrieve
  \item $N_{\text{max}}$: Maximum number of questions per name
  \item $U_{\text{max}}$: Maximum allowed usage per choice
  \item $\mathcal{B}$: Set of banned IDs due to usage limit
  \item $\mathcal{Q}$: Output set of generated multiple-choice VQAs
  \item $\mathcal{C}$: Set of already-used choice combinations (as hashable sets)
\end{itemize}

\subsubsection{Avoid list}

The comprehensive \textit{avoid list} is presented in Figure \ref{fig:avoid_list}. This list has been meticulously compiled based on the insights provided by annotators to prevent overlap between countries for organizing visual options within various question types. It indicates that cultural artifacts positioned within the same row are excluded from the sampling process for visual options. For example, suppose that the correct answer is an image from \textit{Indonesia/music/indonesian sape} in the context of the VQA framework during the initial phase. In that case, images associated with \textit{Malaysia/music/malaysian sape} and \textit{Philippines/music/philipino kudyapi} are systematically excluded from consideration.

\subsubsection{Future work}

We plan to organize a comprehensive challenge inspired by \cite{Damen2022RESCALING, satar2023exploitingsemanticrolecontextualized, bhatia-etal-2024-local_GlobalRG, nayak-etal-2024-benchmarking_CulturalVQA} and also include human evaluations inspired by \cite{Ilaslan_Köksal_Lin_Satar_Shou_Xu_2025} for comparison.

\begin{figure}[!htb]
\centering
  \includegraphics[width=0.92\columnwidth]{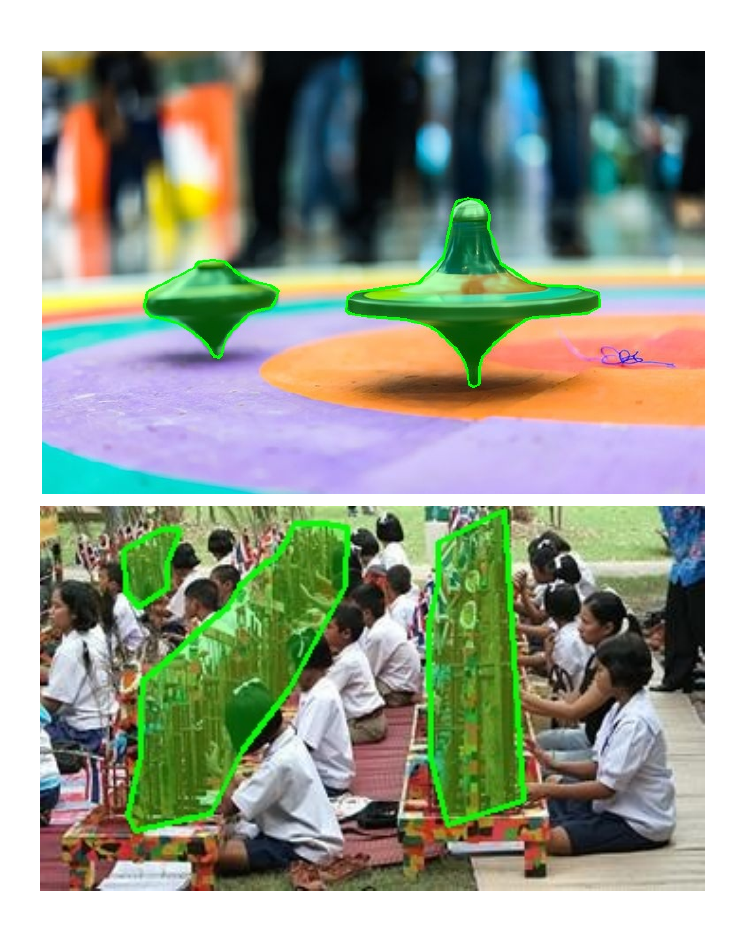}
  \caption{Two images that we eliminated.}
  \label{fig:eliminated_img}
\end{figure}

\begin{figure*}[!htb]
\centering
  \includegraphics[width=1\textwidth]{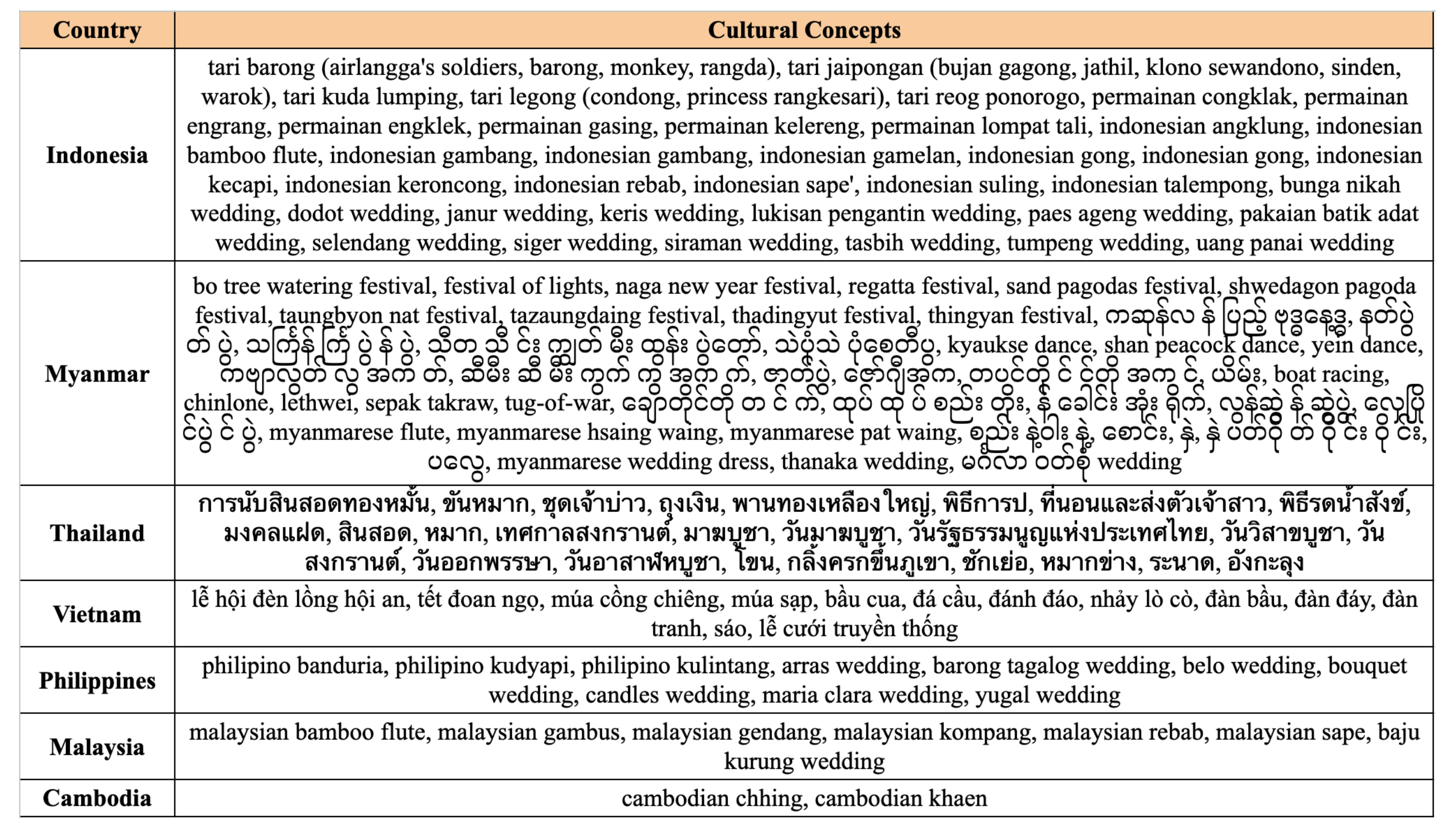}
  \caption{Compilation of cultural concepts addressed in SCB.}
  \label{fig:all_list}
\end{figure*}

\begin{figure*}[!htb]
\centering
  \includegraphics[width=\textwidth]{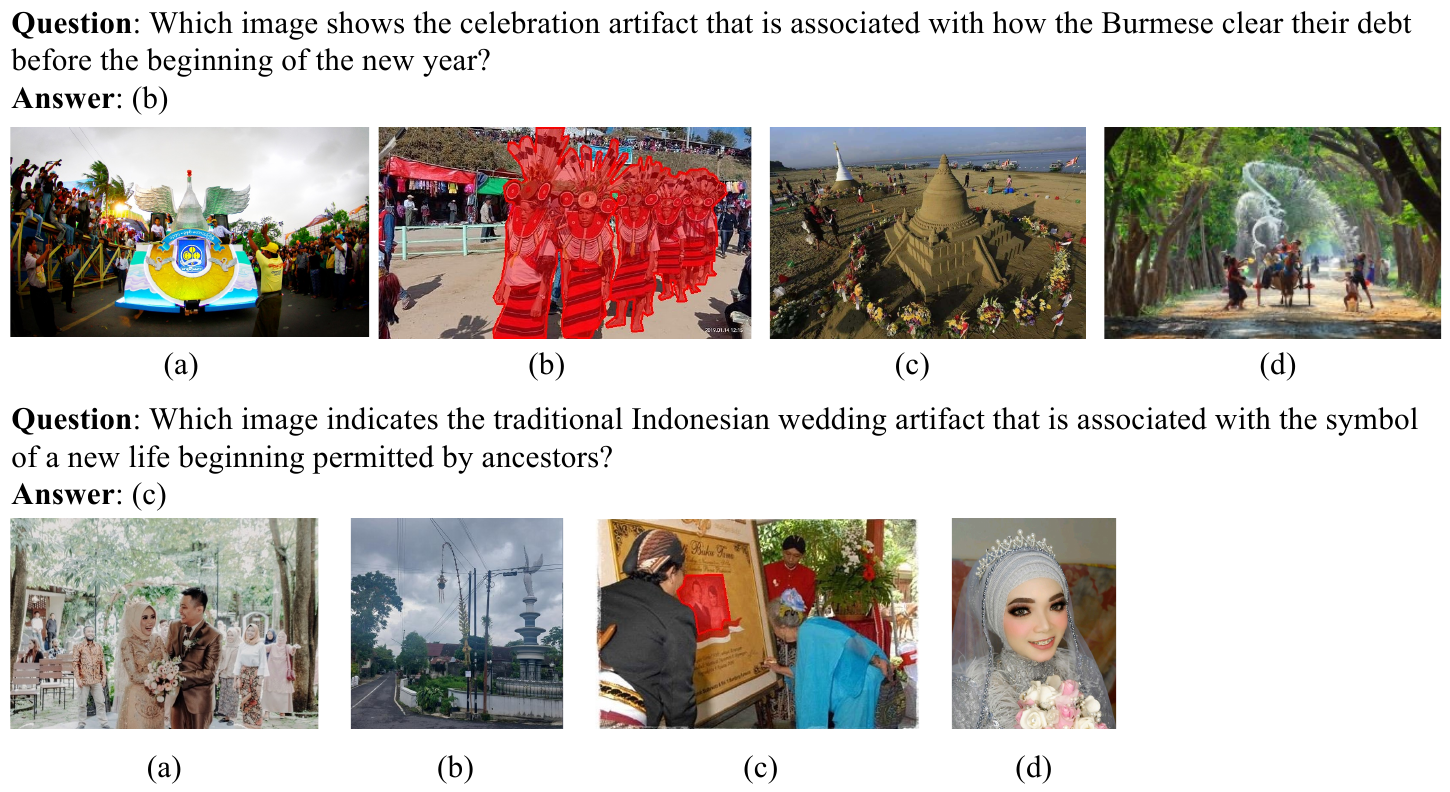}
  \caption{Multiple-choice VQA examples. The red masks in the correct option demonstrate the supporting evidence.}
  \label{fig:mcq_example}
\end{figure*}

\begin{figure*}[!htb]
\centering
  \includegraphics[width=1\textwidth]{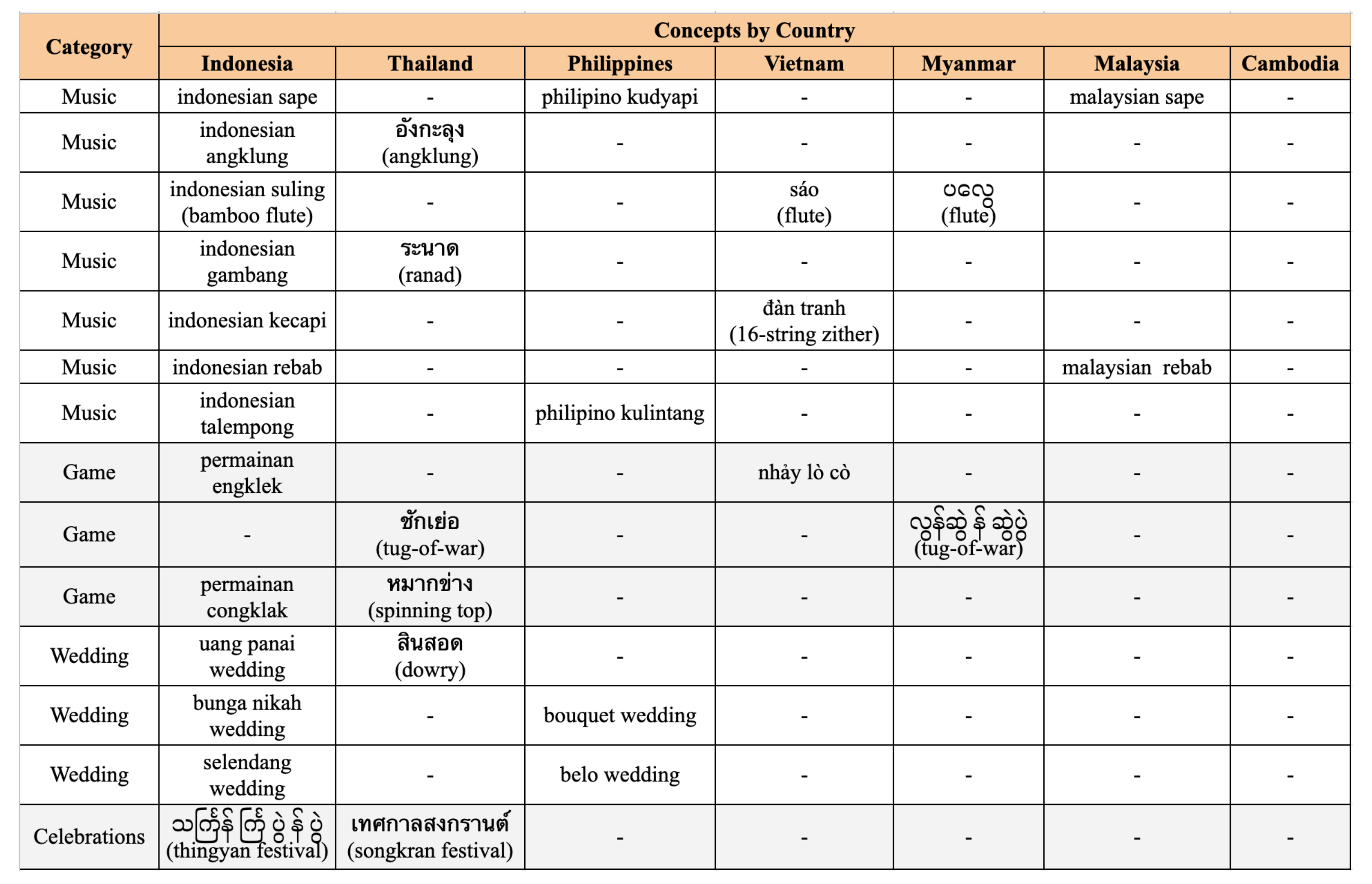}
  \caption{The avoid list for organizing visual options within various question types.}
  \label{fig:avoid_list}
\end{figure*}

\begin{algorithm}[!htb]
\begin{algorithmic}[1]
\STATE Initialize usage counter $\mu : \mathbb{Z} \rightarrow \mathbb{N}$ for all IDs
\STATE Initialize $\mathcal{Q} \leftarrow \emptyset$, $\mathcal{B} \leftarrow \emptyset$, $\mathcal{C} \leftarrow \emptyset$
\FOR{each unique name $n$ in $\mathcal{D}$}
  \STATE Extract $\text{Country}(n), \text{Category}(n)$
  \STATE Let $\mathcal{D}_n \subset \mathcal{D}$ be the $N_{\text{max}}$ samples with name $n$
  \FOR{each sample $q \in \mathcal{D}_n$}
    \STATE Use $\mathcal{V}$ to retrieve top-$k$ similar items $\mathcal{S}$ where $\text{Country}(s) = \text{Country}(n)$, $\text{Category}(s) = \text{Category}(n)$, $\text{Name}(s) \ne n$, and $\text{ID}(s) \notin \mathcal{B}$
    \FOR{each triple $(s_1, s_2, s_3) \subset \mathcal{S}$}
      \IF{each $s_i$ has $\mu(s_i) < U_{\text{max}}$ and $\{\text{ID}(s_i)\} \notin \mathcal{C}$}
        \STATE Form choice set $\mathcal{A} = \{s_1, s_2, s_3, q\}$ with $q$ as the correct answer
        \STATE Add $\text{ID}(\mathcal{A})$ to $\mathcal{C}$, update $\mu$
        \STATE Add $\mathcal{A}$ to $\mathcal{Q}$
        \STATE \textbf{break}
      \ENDIF
    \ENDFOR
    \IF{no valid triple found}
      \STATE Sample 3 random distractors $\mathcal{R}$ satisfying above constraints
      \IF{$|\mathcal{R}| = 3$}
        \STATE Form choice set $\mathcal{A} = \mathcal{R} \cup \{q\}$ and update $\mu$, $\mathcal{C}$
        \STATE Add $\mathcal{A}$ to $\mathcal{Q}$
      \ENDIF
    \ENDIF
  \ENDFOR
\ENDFOR
\RETURN $\mathcal{Q}$
\end{algorithmic}
\caption{Type 1 ($\mathcal{D}, \mathcal{V}, N_{\text{max}}, U_{\text{max}}, k$)}
\label{algo:1}
\end{algorithm}

\begin{algorithm}[!htb]
\begin{algorithmic}[1]
\STATE Initialize usage counter $\mu$, banned ID set $\mathcal{B}$, choice hash set $\mathcal{C}$, and output $\mathcal{Q}$
\FOR{each unique name $n$ in $\mathcal{D}$}
  \STATE Extract $\text{Country}(n), \text{Category}(n)$
  \STATE Let $\mathcal{D}_n \subset \mathcal{D}$ be up to $N_{\text{max}}$ rows with name $n$
  \FOR{each sample $q \in \mathcal{D}_n$}
    \STATE Use $\mathcal{V}$ to retrieve $\mathcal{S}$ where $\text{Country}(s) \ne \text{Country}(n)$, $\text{Category}(s) = \text{Category}(n)$, and $\text{ID}(s) \notin \mathcal{B}$
    \FOR{triplets $(s_1, s_2, s_3)$ with distinct countries}
      \IF{all $\mu(s_i) < U_{\text{max}}$ and $\{\text{ID}(s_i)\} \notin \mathcal{C}$}
        \STATE Form $\mathcal{A} = \{s_1, s_2, s_3, q\}$ with $q$ correct
        \STATE Update $\mu$, $\mathcal{C}$, add $\mathcal{A}$ to $\mathcal{Q}$
        \STATE \textbf{break}
      \ENDIF
    \ENDFOR
    \IF{no valid triplet found}
\STATE Sample $\mathcal{R}$ from $\mathcal{D}$ such that country of $r$ is not equal to country of $n$, \\
      category of $r$ is equal to category of $n$, and name of $r$ is not equal to $n$
      \IF{$|\mathcal{R}| = 3$}
        \STATE Form $\mathcal{A} = \mathcal{R} \cup \{q\}$ and update $\mu$, $\mathcal{C}$
        \STATE Add $\mathcal{A}$ to $\mathcal{Q}$
      \ENDIF
    \ENDIF
  \ENDFOR
\ENDFOR
\RETURN $\mathcal{Q}$
\end{algorithmic}
\caption{Type 2 multiple-choice questions}
\label{algo:2}
\end{algorithm}

\begin{algorithm}[!htb]
\begin{algorithmic}[1]
\STATE Initialize choice usage $\mu$, seen choice sets $\mathcal{C}$, output set $\mathcal{Q}$
\STATE Let $\mathcal{O}$ be original choice sets from $\mathcal{D}$ (to avoid duplicates)
\STATE $\mathcal{C} \gets \mathcal{O}$
\FOR{each question $q \in \mathcal{D}$}
  \STATE Set $used\_choices \gets \emptyset$
  \FOR{$e = 1$ to $E_{\text{max}}$}
    \STATE Extract correct answer $a^*$ with its country, category, and name
    \STATE Extract top distractor $a'$ from $q$ (highest score $\ne -1.0$)
    \STATE Collect banned triples from $a^*$ and $a'$:
    country, category, and name
    \STATE Initialize $choices \gets \{a^*, a'\}$,
    and record used countries and names
    \STATE Let $\mathcal{P} \gets$ opposite type pool
    (type1 if $q$ is type2, else type2)
    \STATE Filter $\mathcal{P}$ to get eligible distractors satisfying:
    same category as $q$, distinct country and name,
    not in banned triples, usage $\mu < U_{\text{max}}$,
    and not in $used\_choices$
    \IF{at least 2 eligible distractors found}
      \STATE Sample 2 distractors $d_1, d_2$ and add to $choices$
      \STATE Update $\mu$ and $used\_choices$
      \STATE Shuffle $choices$ and assign to $q_e$
      \STATE Set correct choice score to $-1.0$,
      others to $-2.0$
      \STATE Mark $q_e.\text{type} \gets$ mixed,
      and update $\mathcal{C}$
      \IF{$choices \notin \mathcal{C}$}
        \STATE Add $q_e$ to $\mathcal{Q}$ and to $\mathcal{C}$
      \ENDIF
    \ENDIF
  \ENDFOR
\ENDFOR
\RETURN $\mathcal{Q}$
\end{algorithmic}
\caption{Type 3 multiple-choice questions}
\label{algo:3}
\end{algorithm}


\end{document}